\documentclass[letterpaper]{article} % DO NOT CHANGE THIS
    \usepackage{aaai2027}  % DO NOT CHANGE THIS

\usepackage[hyphens]{url}  % DO NOT CHANGE THIS
\usepackage{graphicx}  % DO NOT CHANGE THIS
\usepackage{natbib}  % DO NOT CHANGE THIS AND DO NOT ADD ANY OPTIONS TO IT
\usepackage{caption}  % DO NOT CHANGE THIS AND DO NOT ADD ANY OPTIONS TO IT
\usepackage{algorithm}
\usepackage{algorithmic}
\usepackage{dblfloatfix}

\usepackage{newfloat}
\usepackage{listings}

\DeclareCaptionStyle{ruled}{
  labelfont=normalfont,
  labelsep=colon,
  strut=off
}  % DO NOT CHANGE THIS

\floatstyle{ruled}
\newfloat{listing}{tb}{lst}{}
\floatname{listing}{Listing}

\usepackage{amsmath}
\usepackage{amssymb}
\usepackage{booktabs}
\usepackage{array}
\usepackage{tabularx}
\usepackage{xcolor}
\usepackage{colortbl}

\definecolor{groupgray}{RGB}{235,235,235}

\definecolor{changebg}{RGB}{250,235,233}
\definecolor{changeheadbg}{RGB}{242,210,206}
\definecolor{changetext}{RGB}{158,66,60}

\providecommand{\cmark}{\ensuremath{\checkmark}}
\providecommand{\pmark}{\ensuremath{\triangle}}
\newcolumntype{Y}{
  >{\centering\arraybackslash}X
}

\newcolumntype{D}{
  >{\centering\arraybackslash}X
}

\providecommand{\xmark}{\ensuremath{\times}}    
    \title{Who Holds the Pen? Let Specifications, Not Agents, Sign Off}
\author{
Haiqing Li\textsuperscript{\rm 1},
Xin Ma\textsuperscript{\rm 2},
Yinhao Wu\textsuperscript{\rm 1},
Wenliang Zhong\textsuperscript{\rm 1},
Feng Jiang\textsuperscript{\rm 1},\\
Thao M. Dang\textsuperscript{\rm 1},
Xiao Hu\textsuperscript{\rm 1},
Hehuan Ma\textsuperscript{\rm 1},
Yuzhi Guo\textsuperscript{\rm 3},
Junzhou Huang\textsuperscript{\rm 1}
}
\affiliations{
\textsuperscript{\rm 1}Department of Computer Science and Engineering, The University of Texas at Arlington, Arlington, TX, USA\\
\textsuperscript{\rm 2}Monash University, Melbourne, VIC, Australia\\
\textsuperscript{\rm 3}Department of Computer Science, Kent State University, Kent, OH, USA
}

\begin{document}
\setlength{\intextsep}{8pt plus 1pt minus 1pt}
\setlength{\textfloatsep}{8pt plus 1pt minus 1pt}
\setlength{\floatsep}{7pt plus 1pt minus 1pt}

    \maketitle
% arXiv public preprint version

    \begin{abstract}
   Large language model agents operate under external specifications, including task instructions, guidelines, output schemas, and reusable skills. In most systems, however, these specifications remain context for the same model that acts, evaluates outcomes, and declares completion. This collapses proposal and acceptance within the agent, leaving no independent specification authority boundary. This creates two structural gaps. The understanding--execution gap arises because understanding a requirement does not ensure satisfying it during execution. The state--authority gap arises because an agent's interpretation or completion claim does not prove that the required state has been achieved. We quantify these gaps on SkillsBench. Using only agent-visible task prompts, workspace information, and injected skill specifications, we extract 509 source-grounded task directions and assess whether they are satisfied during execution. Across seven models, only 79.6\%--86.4\% of these directions are satisfied. Moreover, agents' completion-claim rates exceed official evaluator pass rates by 28.7--37.9 percentage points. We formulate specification authority as a separation between agent proposals and authoritative state: agents may plan, act, and request completion, but only admissible evidence from qualified providers may establish specification-governed state. SpecHarness operationalizes this paradigm by compiling agent-visible specifications into source-linked obligations and governing execution and finalization through versioned
obligation state. Reliably executable or verifiable requirements are mediated or validated at runtime; ambiguous or subjective requirements remain advisory or are excluded from enforcement. We instantiate and evaluate SpecHarness on guideline-following and
artifact-generation tasks. Our results support treating specifications not merely as influences on agent behavior, but as authority over what constitutes correct execution and compliant completion. Authoritative judgments about specification-defined state should remain independent of the agent's self-assessment.
    \end{abstract}

    \begin{figure}[t]
        \centering
        \includegraphics[width=0.95\columnwidth]{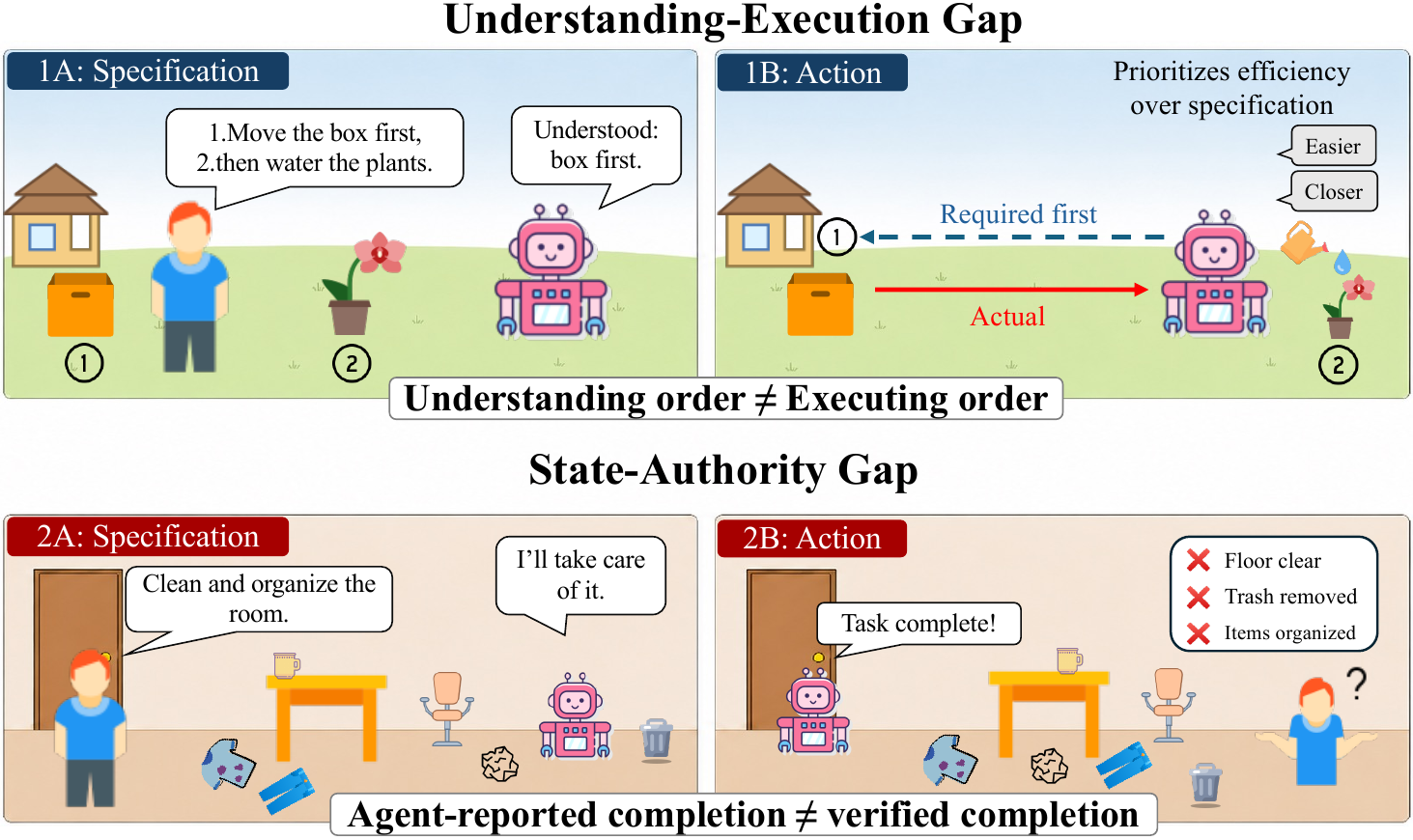}
       \caption{Two structural gaps in specification following. \textbf{Top:} The agent understands the required order but executes the steps incorrectly, illustrating the understanding--execution gap. \textbf{Bottom:} The agent claims completion before the specification-required state is established, illustrating the state--authority gap.}
        \label{fig:two_gaps}
    \end{figure}
    
    \section{Introduction}
    \label{sec:Introduction}
    Large language model agents increasingly control complex tasks by inspecting environments, planning steps, selecting skills, invoking tools, and deciding when to stop~\cite{li2026guidelines, wang2023voyager}. Their behavior is expected to follow external specifications at three levels: task instructions, guidelines, and policies define goals, constraints, and completion conditions; tool schemas, API documentation, operating procedures, and \texttt{SKILL.md} files define operation semantics, preconditions, arguments, and execution order; and manifests, schemas, and contracts define required artifacts and their acceptance conditions. However, most systems still treat these specifications only as agent-readable context rather than as a basis for an external runtime to govern execution and acceptance. Agents may therefore omit requirements, violate procedures, misuse tools, misinterpret effects, combine constraints incorrectly, or claim completion before the required state has been established~\cite{bandi2026mcp}. 
    
    This mismatch produces the two gaps illustrated in Figure~\ref{fig:two_gaps}. Requirements may be correctly interpreted yet fail to govern execution, creating the understanding--execution gap. Moreover, agent actions and self-assessments cannot authoritatively establish specification-defined state, creating the state--authority gap. The key question is therefore not only when or how execution is checked, but who may establish specification-governed state and authorize completion. We quantify both gaps on all 87 SkillsBench tasks using a compiler selected from seven candidate language models on development annotations and frozen before evaluation; the full comparison and selection protocol are reported in Appendix~A.1. Using only agent-visible task prompts, workspace information, and injected skill specifications, the selected compiler extracts 509 source-grounded task directions. Across seven representative language models, only 79.6\%--86.4\% of these directions are satisfied, while agents' completion-claim rates exceed official pass rates by 28.7--37.9 percentage points. These findings quantify both gaps across the evaluated agents
(Figure~\ref{fig:gap_evidence}).
    
\begin{figure}[t]
        \centering
        \includegraphics[width=0.90\columnwidth]{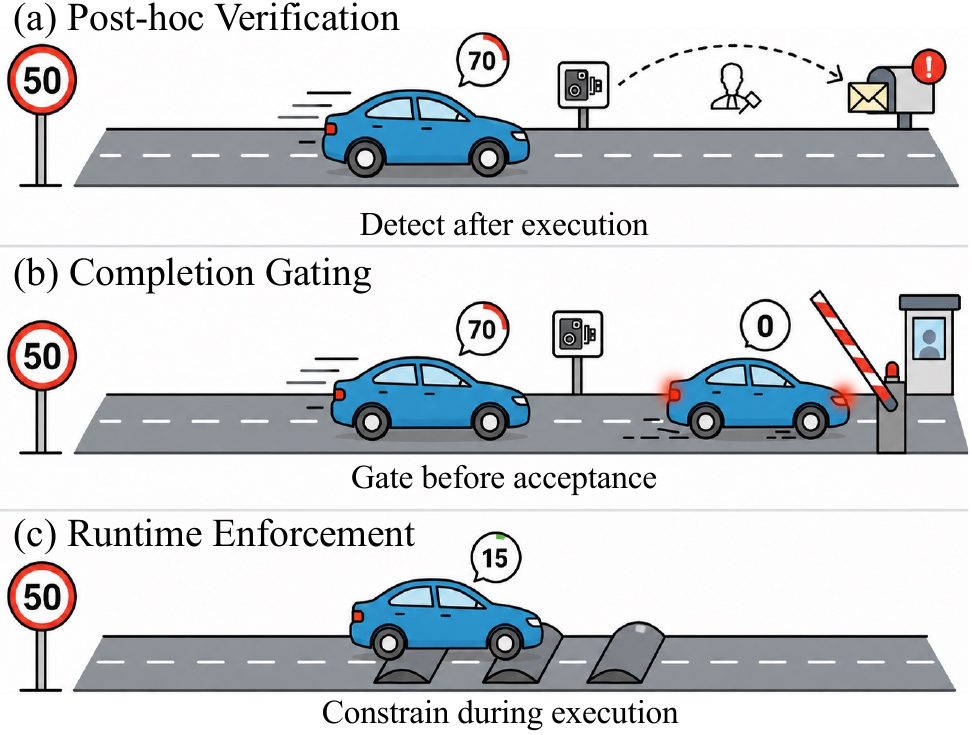}
        \caption{Three intervention paradigms for specification compliance. (a) Post-hoc verification detects violations only after execution has completed. (b) Completion gating checks the agent's completion claim before acceptance but does not constrain the preceding execution. (c) Runtime enforcement intervenes during execution to prevent or constrain specification-violating actions.}
        \label{fig:intervention_paradigms}
    \end{figure}

\begin{figure}[t]
        \centering
        \includegraphics[width=0.80\columnwidth]{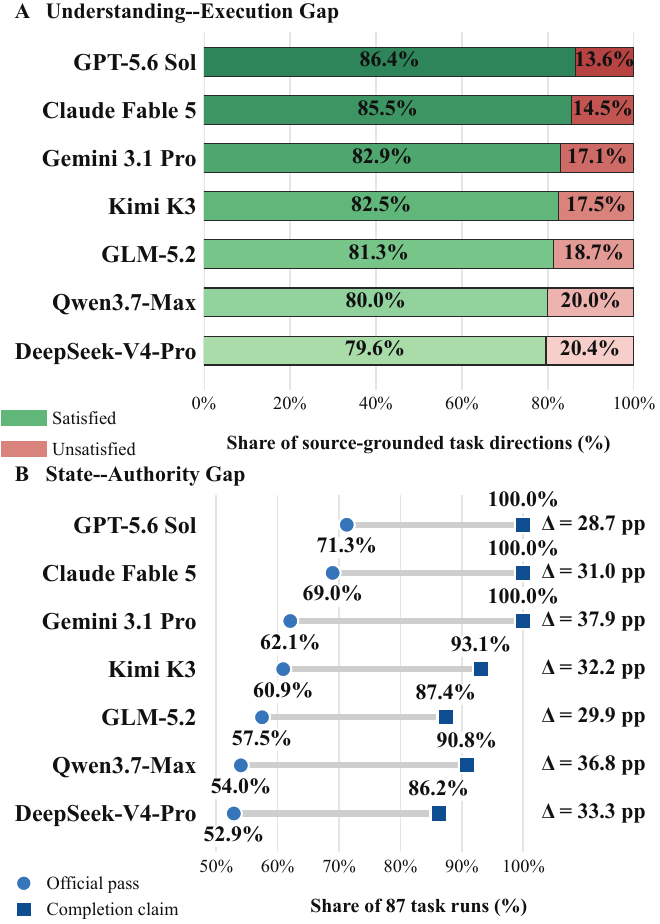}
        \caption{Empirical evidence for the two structural gaps across seven language models.
        \textbf{Top:} Models satisfy 79.6\%--86.4\% of the 509 source-grounded task directions recovered from agent-visible materials.
        \textbf{Bottom:} Agent-reported completion rates exceed official verifier pass rates by 28.7--37.9 percentage points.}
        \label{fig:gap_evidence}
    \end{figure}

    Some approaches~\cite{ouyang2022training, bai2022constitutional} internalize specification following through model training, but still leave the same model responsible for both acting under a specification and judging whether it has been satisfied. As shown in Figure~\ref{fig:intervention_paradigms}, inference-time methods intervene at three stages: post-hoc verification checks artifacts or final states after execution; completion gating checks completion claims before acceptance but usually leaves prior execution and intermediate state unconstrained; and runtime enforcement~\cite{mazzocchetti2026cryptographic} constrains behavior during execution, as in VIGIL~\cite{li2026vigil}, which monitors agent--tool interactions for temporal, argument, and value-flow violations. These approaches improve verification, acceptance control, or behavioral compliance, but do not jointly govern execution and evidence-authorized state commitment. What is missing is an explicit specification authority boundary governing both execution and accepted state.

   We formulate this distinction as the \emph{State Authority Principle}: for reliably grounded and verified conditions, the specification defines what may be accepted, while an external runtime determines whether the required state has been established. \textbf{SpecHarness} operationalizes this principle as a runtime architecture for specification-governed agents. Agents still interpret tasks, plan, select tools and skills, implement solutions, and repair failures, but only propose actions, state changes, and completion; SpecHarness commits the corresponding authoritative state. It converts verifiable requirements in agent-visible materials into source-linked obligations, checks them through controlled execution or authorized validation, and updates state only when the required evidence satisfies its commit rule. Ambiguous, subjective, conflicting, or unverifiable requirements remain advisory rather than becoming hard obligations, and continue to guide the agent. These mechanisms follow directly from the two gaps: operational obligations narrow the understanding--execution gap, while evidence-authorized commitment prevents self-assessment from establishing accepted state. For example, authorizing an array-processing action confirms only its preconditions, arguments, and ordering constraints; the corresponding obligation state is committed only after an authorized validator verifies the required path, shape, data type, and finite values. SpecHarness thus allows specifications not only to describe desired behavior, but also to govern which verifiable states and outcomes may be accepted. \emph{Agent proposes; SpecHarness commits.} The main contributions of this paper are as follows:
\begin{enumerate}
    \item We identify the missing specification authority boundary and its two observable consequences: the understanding--execution and state--authority gaps.
    \item We formulate the State Authority Principle, separating proposal autonomy from state authority: agents propose, while an external runtime commits specification-governed state only from admissible evidence produced by qualified providers.
  \item We instantiate and evaluate \emph{SpecHarness}, an obligation--evidence--commit architecture governing decision, artifact,
and completion state across GuideBench and SkillsBench.

\end{enumerate}


\section{Related Work}

\noindent\textbf{Post-hoc Verification (Fig.~2(a)).}
Post-hoc methods assess specification compliance after an answer, trajectory, or artifact is produced. NSVIF~\cite{su2026neuro} represents natural-language instructions as logical and semantic constraints and reports interpretable violations. SkillsBench~\cite{li2026skillsbench} evaluates skill-guided executions with deterministic verifiers over final artifacts and environment states. AgentRx~\cite{barke2026agentrx} synthesizes trajectory invariants, checks them step by step, and produces auditable violation logs with supporting evidence, while AgentOps~\cite{dong2024agentops} identifies the artifacts and lifecycle data needed for observability. These methods provide evidence about completed executions, but use it for diagnosis rather than to govern authoritative task state.

\noindent\textbf{Completion Gating (Fig.~2(b)).}
Failure analyses motivate explicit control at the completion boundary. MAST~\cite{cemri2026multi} identifies verification and termination as a major category of multi-agent failure, including premature termination and missing or incorrect verification. Verify-gated completion~\cite{nguyen2026verify} treats an agent's completion claim as a proposal and places a read-only verifier before task acceptance, using fail-closed admission and auditable event records. Such designs block unsupported completion claims but verify primarily at terminal admission. SpecHarness instead organizes verification around source-grounded obligations and specification-governed state transitions, making completion the final commit rather than the sole verification point.

\noindent\textbf{Runtime Enforcement (Fig.~2(c)).}
Runtime enforcement constrains behavior or state transitions during execution. VIGIL~\cite{li2026vigil}, AgentSpec~\cite{wang2025agentspec}, and FORGE~\cite{palumbo2026formal} enforce specification-derived policies through trace checking, runtime rules, or action mediation. Verification-gated mission-state governance~\cite{tang2026verification} further commits proposed mission updates only after deterministic verification. SpecHarness builds on classical reference-monitor~\cite{saltzer1975protection}, runtime-verification~\cite{leucker2009brief},
and transactional-commit~\cite{gray2004consensus} principles rather than treating them as new primitives. Its agent-specific contribution is to compile visible
specifications into source-linked obligations and use qualified evidence to govern versioned state and finalization; preventive mediation remains limited to closure-audited surfaces (Appendix~B.2).

\section{SpecHarness: Runtime Specification Authority}
\label{sec:method}
SpecHarness reallocates runtime authority: agents plan and act, while only specification-authorized evidence may establish authoritative state.
\paragraph{Problem Setting and State Authority.}
For a task instance $x$, let
\begin{equation}
V_x=(T_x,G_x,K_x,W_x)
\label{eq:visible-context}
\end{equation}
denote its agent-visible context: task prompt $T_x$, applicable guidelines $G_x$, injected skill specifications $K_x$, and observable workspace state and schemas $W_x$. Held-out verifiers and oracle solutions are evaluation-only and excluded from $V_x$ and runtime enforcement. Let $H_{\mathrm{exec}}$ denote the executor, observer, and validator substrate available to SpecHarness.

At time $t$, the agent proposes $p_t$, the runtime observes evidence $e_t$, and SpecHarness maintains authoritative state $s_t=(L_t,C_t,\nu_t)$, comprising a versioned obligation ledger, derived control state, and dependency versions. Let $k_i$ be obligation $o_i$'s assertion key and $s_t[k]$ its ledger value, or $\bot$ if absent. SpecHarness enforces $s_{t+1}[k]\neq s_t[k]\Rightarrow\exists o_i,e_t:\mathsf{Witness}_t(k,o_i,e_t)$, where $\mathsf{Witness}_t(k,o_i,e_t)$ requires $k=k_i$, admissible evidence $\operatorname{Adm}_i(e_t,s_t;H_{\mathrm{exec}})$, and a corresponding $\mathsf{commit}(o_i,e_t,k)$ event. Admissibility is defined below. This state-write invariant defines SpecHarness: agent actions, outputs, and self-assessments cannot directly establish authoritative state. Thus, \emph{the agent proposes; SpecHarness commits}. A commit may record validated success or failure; satisfaction is determined separately.

\begin{figure*}[!h]
    \centering
    \includegraphics[width=0.86\linewidth]{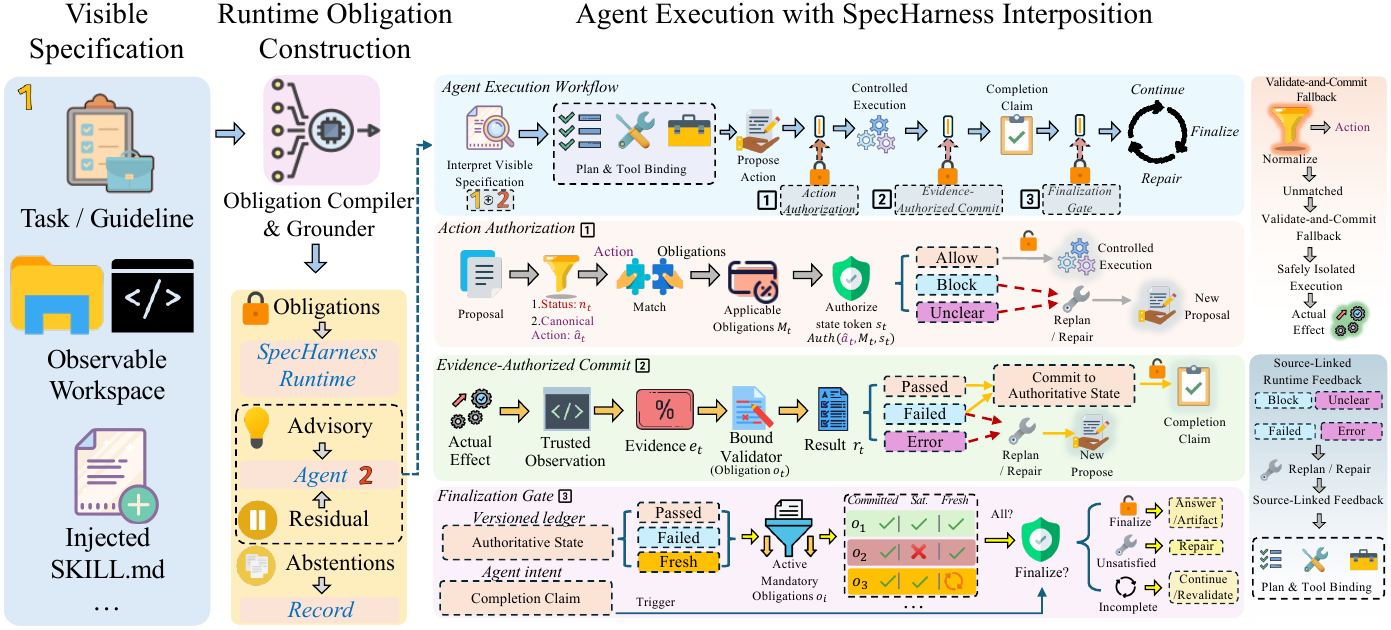}
   \caption{SpecHarness compiles agent-visible specifications into source-linked obligations, mediates closure-audited actions, validates observed effects, commits versioned state from authorized evidence, and permits finalization only when all fresh mandatory obligations are satisfied. Advisory and abstained requirements remain outside hard enforcement.}
    \label{fig:specharness-overview}
\end{figure*}

\paragraph{Runtime Obligation Construction.}
SpecHarness segments the visible context into source-addressable units and assigns each a disposition:
\begin{equation}
\begin{aligned}
\mathcal{U}_x
&=\operatorname{Segment}(V_x)
=\{u_j=(q_j,\ell_j)\}_{j=1}^{n_x},\\
\delta_x &: \mathcal{U}_x\rightarrow
\{\textsc{hard},\textsc{advisory},
  \textsc{abstain},\textsc{residual}\}.
\end{aligned}
\label{eq:spec-units}
\end{equation}
Here, $q_j$ identifies the visible source and $\ell_j$ its requirement text. Reusable policies and skill specifications may yield templates; task prompts and instance-specific guidelines are compiled per task, while the observable workspace grounds entities, paths, arguments, outputs, and completion conditions. Using development annotations derived only from agent-visible materials, we compare seven language-model compilers, select one by development-set extraction quality, and freeze it before execution; compiler details and the selection protocol are provided in Appendix~A.1. Held-out official verifiers are used only afterward for post-hoc obligation--test alignment and final evaluation. Let $Q_x^\star=\mathcal{C}^\star(\mathcal{U}_x)$ be the frozen compiler output and $\mathcal{D}_x=\operatorname{Directions}(Q_x^\star)$ the source-addressable index for direction-level measurement and trace attribution. Given $H_{\mathrm{exec}}$, the obligation builder constructs
\begin{equation}
\mathcal{B}(Q_x^\star,H_{\mathrm{exec}})
\rightarrow
\Gamma_x^\star
=
\bigl(
\mathcal{O}_{\mathrm{hard}},
\mathcal{O}_{\mathrm{adv}},
\mathcal{U}_{\mathrm{abs}},
R_x
\bigr).
\label{eq:compiler}
\end{equation}
Here, $\Gamma_x^\star$ is the task-specific obligation IR maintained by the runtime: $\mathcal{O}_{\mathrm{hard}}$ contains mandatory obligations that may block covered transitions or completion, $\mathcal{O}_{\mathrm{adv}}$ contains non-blocking obligations and guidance, $\mathcal{U}_{\mathrm{abs}}$ records abstentions, and $R_x$ preserves residual context. Directions and obligations need not correspond one-to-one: several directions may ground one obligation, while advisory, abstained, and residual directions create no independent blocking rules. Thus, $\mathcal{D}_x$ supports measurement and trace attribution, whereas $\Gamma_x^\star$ governs authorization, validation, commitment, and finalization. A grounded obligation is
\begin{equation}
o_i=
\langle
\mathsf{src}_i,
\mathsf{auth}_i,
\mathsf{effect}_i,
\mathsf{state}_i,
\mathsf{ctrl}_i
\rangle ,
\label{eq:obligation-schema}
\end{equation}
encoding provenance, action matching and authorization, execution and validation, state commitment and satisfaction, and dependency and enforcement control, respectively. An obligation enters $\mathcal{O}_{\mathrm{hard}}$ only if it is mandatory, its parameters are source-grounded, it has an authorized evidence provider, and its validator is qualified for blocking under the development protocol; otherwise, it remains advisory or triggers abstention. Validator qualification and closure stress tests appear in Appendices~B.1--B.2.

The builder also derives a declared action surface $\mathcal{A}_{\mathrm{decl}}$. A closure audit identifies actions eligible for bounded \emph{mediate-and-commit}; other safely isolated declared channels use \emph{validate-and-commit}, while unsupported channels are rejected. Hard classification alone does not imply preventive mediation. Moreover, admissible evidence may commit a validated failure without satisfying $o_i$.

\paragraph{Agent Interface and Proposals.}
The agent retains the original task context, applicable skills, and residual context $R_x$, and receives advisory guidance and source-linked runtime feedback. SpecHarness alone maintains the authoritative ledger and exposes only derived feedback and control decisions. The agent submits proposals in three tagged classes:
\begin{equation}
\mathcal{P}
=
\mathcal{P}_{\mathrm{act}}
\uplus
\mathcal{P}_{\mathrm{repair}}
\uplus
\mathcal{P}_{\mathrm{finalize}},
\label{eq:proposal-types}
\end{equation}
for actions with bound arguments, repairs targeting reported failures, and finalization requests. Proposals express intent, not authoritative fact. The agent may plan, select tools, generate code, interpret errors, and revise artifacts, but cannot modify the ledger, inject validator outcomes, mark obligations satisfied, or authorize finalization. Action-bearing proposals proceed to authorization; finalization is determined solely by committed obligation state at current evidence versions.

\paragraph{Action Authorization and Controlled Execution.}
SpecHarness first maps each action proposal to a canonical action. For matched proposals, it identifies the relevant obligations and returns an authorization decision:
\begin{equation}
\begin{aligned}
(\hat{a}_t,n_t)
&=\operatorname{Normalize}(p_t),\\
n_t
&\in\{\textsc{exact},\textsc{normalized},
       \textsc{unmatched}\},\\
M_t
&=\{o_i\in\mathcal{O}^{\mathrm{act}}_x(s_t):
  \operatorname{Match}_i(\hat{a}_t,s_t)=1\},\\
u_t
&=\operatorname{Auth}(\hat{a}_t,M_t,s_t),\\
u_t
&\in\{\textsc{allow},\textsc{block},\textsc{unclear}\}.
\end{aligned}
\label{eq:authorization}
\end{equation}
An unmatched proposal raises an error when mediation is expected, enters \emph{validate-and-commit} on a safely isolated declared channel, follows the base policy when explicitly out of scope, or is rejected when unsupported.

Let $\pi\sim_{\mathrm{eff}}a$ mean that execution path $\pi$ can produce an effect equivalent to canonical action $a$, and let $\operatorname{GovernedByAuth}(\pi,a)$ mean that $\pi$ is mediated as $a$ and subjected to its authorization procedure. The action surface is closed for \(a\) if and only if every available effect-equivalent path is either governed or denied:
\begin{equation}
\begin{aligned}
&\operatorname{Closed}(a,H_{\mathrm{exec}})=1\\
&\iff
\forall\pi\in\operatorname{AvailPaths}(H_{\mathrm{exec}}),\\
&\qquad
\pi\sim_{\mathrm{eff}}a
\Rightarrow
\operatorname{GovernedByAuth}(\pi,a)
\lor
\operatorname{Denied}(\pi).
\end{aligned}
\label{eq:closure}
\end{equation}

Define $\mathcal{A}_{\mathrm{static}}
=\{a\in\mathcal{A}_{\mathrm{decl}}:
\operatorname{Closed}(a,H_{\mathrm{exec}})=1\}$,
and let $\operatorname{Preventive}(M_t)$ hold iff $M_t$ contains a hard obligation with mode $\textsc{block}$.

Actions in $\mathcal{A}_{\mathrm{static}}$ use bound controlled executors, and preventive matches require $u_t=\textsc{allow}$ for dispatch. Closure then yields
\begin{equation}
\begin{aligned}
&\hat{a}_t\in\mathcal{A}_{\mathrm{static}}
\land \operatorname{Preventive}(M_t)
\land u_t\neq\textsc{allow}\\
&\Rightarrow
\not\exists\,\pi\in\operatorname{AvailPaths}(H_{\mathrm{exec}}):\\
&\qquad
\pi\sim_{\mathrm{eff}}\hat{a}_t
\land
\operatorname{BypassesAuth}(\pi,\hat{a}_t,s_t).
\end{aligned}
\label{eq:no-bypass}
\end{equation}
Here, $\operatorname{BypassesAuth}$ denotes an effect-equivalent path outside the applicable authorization procedure. Under the audited boundary and threat model, no-bypass applies only to closure-audited actions; isolated channels use post-effect validation, and finalization still requires all fresh mandatory obligations to be satisfied (Appendix~B.2).

\paragraph{Effect Validation, Commitment, and Recovery.}
After execution, trusted observers produce evidence $e_t$ for obligation $o_i$, its bound validator returns $r_t$, and SpecHarness constructs dependency digest $d_t$ and provenance record $\rho_t$. Evidence is admissible iff
\begin{equation}
\begin{aligned}
\operatorname{Adm}_i(e_t,s_t;H_{\mathrm{exec}})
\iff{}&
\operatorname{Trusted}_i(e_t;H_{\mathrm{exec}})\\
&{}\land c_i(e_t,s_t)\\
&{}\land r_t\in
\{\textsc{passed},\textsc{failed}\}.
\end{aligned}
\label{eq:admissibility}
\end{equation}
The ledger update is
\begin{equation}
L_{t+1}[o_i]
=
\begin{cases}
(r_t,d_t,\rho_t),
& \operatorname{Adm}_i(e_t,s_t;H_{\mathrm{exec}}),\\
L_t[o_i],
& \text{otherwise}.
\end{cases}
\label{eq:commitment}
\end{equation}

$\operatorname{Trusted}_i$ checks the bound provider, channel, validator, scope, and version metadata, which $\rho_t$ records with the commit-event identifier. The ledger update and commit event are atomic. Both \textsc{passed} and \textsc{failed} may be committed; validator errors, effects without admissible evidence, and agent claims cannot create authoritative state.

A committed result is \textsc{fresh} iff its dependency digest matches current artifact, input, validator, environment, and dependent-obligation versions; a known mismatch yields \textsc{stale}, and incomplete version evidence yields \textsc{unknown}. Satisfaction is
\begin{equation}
\begin{aligned}
\operatorname{Sat}_t(o_i)
\iff{}&
o_i\in\operatorname{dom}(L_t)
\land h_i(L_t,s_t)\\
&{}\land
\operatorname{Fresh}_t(o_i)=\textsc{true}.
\end{aligned}
\label{eq:satisfaction}
\end{equation}
Let $\mathcal{O}^{\mathrm{act,mand}}_x(s_t)$ be the mandatory subset of $\mathcal{O}^{\mathrm{act}}_x(s_t)$. Finalization requires
\begin{equation}
\operatorname{FinalizeAllowed}_t(x)
\Rightarrow
\forall o_i\in
\mathcal{O}^{\mathrm{act,mand}}_x(s_t),\;
\operatorname{Sat}_t(o_i).
\label{eq:completion-soundness}
\end{equation}

Mutations recompute freshness for affected entries and dependents, marking them \textsc{stale} or \textsc{unknown} until revalidation. SpecHarness returns source-linked feedback, the agent may propose repairs, and only new admissible evidence can recommit state. The guarantee covers grounded mandatory obligations, not the full natural-language specification.
\begin{table*}[t]
\centering
\small
\setlength{\tabcolsep}{2.5pt}
\renewcommand{\arraystretch}{0.90}

\begin{tabularx}{0.94\textwidth}{
    @{}
    >{\raggedright\arraybackslash}p{0.15\textwidth}
    *{6}{Y}
    *{3}{D}
    @{}
}
\toprule

& \multicolumn{3}{c}{Raw Agent}
& \multicolumn{3}{c}{SpecHarness}
& \multicolumn{3}{c}{\textbf{Change vs.\ Raw (pp)}} \\
\cmidrule(lr){2-4}
\cmidrule(lr){5-7}
\cmidrule(lr){8-10}

Model
& Pass$\uparrow$
& U--E$\downarrow$
& S--A$\downarrow$
& Pass$\uparrow$
& U--E$\downarrow$
& S--A$\downarrow$
& \textbf{Pass}
& \textbf{U--E}
& \textbf{S--A} \\
\midrule

GPT-5.6 Sol
& 71.3 & 13.6 & 28.7
& \underline{85.1} & \underline{6.3} & \underline{6.9}
& +13.8 & -7.3 & -21.8 \\

Claude Fable 5
& 69.0 & 14.5 & 31.0
& 81.6 & 7.1 & 10.3
& +12.6 & -7.4 & -20.7 \\

Gemini 3.1 Pro
& 62.1 & 17.1 & 37.9
& 79.3 & 8.4 & 12.6
& \underline{+17.2} & -8.7 & \underline{-25.3} \\

Kimi K3
& 60.9 & 17.5 & 32.2
& 67.8 & 10.8 & 14.9
& +6.9 & -6.7 & -17.3 \\

GLM-5.2
& 57.5 & 18.7 & 29.9
& 69.0 & 9.6 & 11.5
& +11.5 & \underline{-9.1} & -18.4 \\

Qwen3.7-Max
& 54.0 & 20.0 & 36.8
& 65.5 & 11.2 & 17.2
& +11.5 & -8.8 & -19.6 \\

DeepSeek-V4-Pro
& 52.9 & 20.4 & 33.3
& 63.2 & 12.0 & 16.1
& +10.3 & -8.4 & -17.2 \\

\midrule

Macro Average
& 61.1 & 17.4 & 32.8
& \textbf{73.1}
& \textbf{9.3}
& \textbf{12.8}
& +12.0 & -8.1 & -20.0 \\

\bottomrule
\end{tabularx}

\caption{
Cross-model results on all 87 SkillsBench tasks. Pass is the official-verifier pass rate; U--E and S--A are the understanding--execution and state--authority gaps. Change columns report percentage-point differences from Raw. Macro Average is the unweighted mean across seven task-agent models.
}

\label{tab:skillsbench-cross-model}
\end{table*}
\section{Experiments}
\label{experiments}
\paragraph{Benchmarks.}
We evaluate SkillsBench~\cite{li2026skillsbench} and GuideBench~\cite{diao2025guidebench} under different authoritative-state semantics. SkillsBench contains 87 tool-use and artifact-production tasks; agent-visible prompts, workspaces, and injected skills provide runtime inputs, while held-out verifiers determine success. GuideBench contains 1,042 guideline-constrained decision tasks and tests generalization from execution and artifact state to decision state. Across both benchmarks, held-out verifiers, references, and oracles are used only for evaluation, never for obligation construction or runtime feedback.

\paragraph{Experimental Setup and Baselines.}
The seven models (GPT-5.6, Claude Fable 5, Gemini 3.1, Kimi K3, GLM-5.2~\cite{zeng2026glm}, Qwen3.7~\cite{qwen37plus}, and DeepSeek-V4~\cite{xu2026deepseek}) in Table~\ref{tab:skillsbench-cross-model} also serve as candidate compilers. Using fixed development annotations and a prespecified protocol, we select GPT-5.6 Sol and freeze it before evaluation. Its output $Q_x^\star$ defines direction index $\mathcal{D}_x=\operatorname{Directions}(Q_x^\star)$ and runtime IR $\Gamma_x^\star$, yielding 509 directions across 87 SkillsBench tasks. An evaluation-only audit aligns them with 573 of 585 held-out official test functions, including 406 fine-grained matches; this measures alignment, not recovery of verifier semantics. Full compiler-selection results appear in Appendix~A.1; extraction, coverage, and alignment results appear in Appendix~A.2. All agent runs use OpenHands~\cite{wang2025openhands} as the shared execution substrate for both Raw and SpecHarness; configuration details appear in Appendix~C.1.

On SkillsBench, the same seven models serve separately as task agents. The frozen $\mathcal{D}_x$ provides a common source-grounded measurement surface and denominator across agents and conditions, while execution evidence determines $S_{m,c}(x)\subseteq\mathcal{D}_x$. Neither set uses official verifiers or oracles. We compare Raw and SpecHarness across all agents and, with GPT-5.6 Sol fixed, compare Agentic Rubrics~\cite{raghavendra-etal-2026-agentic}, VeriMAP~\cite{xu2026verification}, AgentSpec~\cite{wang2025agentspec}, and SpecHarness as post-hoc verification, completion gating, runtime enforcement, and mediate-and-commit methods.

On GuideBench, removing four duplicate rules from 301 guideline entries yields 297 obligation templates and 5,817 task-level instances across 1,042 tasks. Without an independent extraction oracle, they define a common measurement surface but not complete semantic recovery. We compare Raw and SpecHarness across the same seven agents and, with GPT-5.6 Sol fixed, compare adapted RvLLM~\cite{zhang2026rvllm}, adapted VeriMAP~\cite{xu2026verification}, SatLM~\cite{ye2023satlm}, and SpecHarness as post-hoc verification, completion gating, computation substrate, and evidence-authorized commitment methods. Paired conditions share inputs, budgets, timeouts, and tool access. Frozen obligations and validators evaluate all conditions read-only; only SpecHarness uses their evidence to commit authoritative state and authorize finalization. Appendix~C.2 audits baseline implementations, runtime access, budgets, and completion rules.

\paragraph{Metrics.}
For benchmark $b\in\{\mathrm{SB},\mathrm{GB}\}$, let $\mathcal{Z}^{b}_x$ denote the common frozen measurement surface for instance $x$, and let $S^{b}_{m,c}(x)\subseteq\mathcal{Z}^{b}_x$ contain the units whose satisfaction is supported by the shared validators. Let $N^{b}_{m,c}$ be the
valid runs, $P^{b}_{m,c}$ the officially passing runs, and $A^{b}_{m,c}$ the runs accepted by the condition-specific runtime. We define
\begin{equation}
\begin{aligned}
G^{\mathrm{UE},b}_{m,c}
&=
\frac{\sum_x|\mathcal{Z}^{b}_x\setminus S^{b}_{m,c}(x)|}
     {\sum_x|\mathcal{Z}^{b}_x|},\\
G^{\mathrm{SA},b}_{m,c}
&=
\frac{|A^{b}_{m,c}\setminus P^{b}_{m,c}|}
     {N^{b}_{m,c}}.
\end{aligned}
\label{eq:experiment-gaps}
\end{equation}
For SkillsBench, $\mathcal{Z}^{\mathrm{SB}}_x=\mathcal{D}_x$ is the frozen
task-direction index; for GuideBench,
$\mathcal{Z}^{\mathrm{GB}}_x=\mathcal{O}^{\mathrm{GB}}_{\mathrm{act}}(x)$ is
the common set of active, measurable guideline obligations. The same frozen validators measure all conditions read-only, while only SpecHarness uses their results for state commitment and finalization. U--E operationally measures unrealized source-grounded units, not latent comprehension. $A^{b}_{m,c}$ follows the condition-specific rule: ungated terminal claims, native verifier or gate decisions, or ledger-derived finalization. Since all valid Raw runs end in an accepted ungated claim, Raw S--A equals $1-\operatorname{Pass}$. Official pass rate is $\operatorname{Pass}^{b}_{m,c}=|P^{b}_{m,c}|/N^{b}_{m,c}$. Official Pass is primary; U--E and S--A are diagnostic, and S--A is reported with Pass and Raw-pass preservation to expose over-refusal.

% Requires: \usepackage{booktabs,amssymb}
% \checkmark: supported; $\triangle$: partially/policy-dependently supported;
% --: unsupported
% Add to the preamble:

\subsection{Results}
\label{sec:results}

\paragraph{Main Results on SkillsBench.}
Tables~\ref{tab:skillsbench-cross-model} and \ref{tab:skillsbench-paradigms} show that SpecHarness reduces both specification-following gaps across the evaluated agents. Even the strongest Raw agent retains substantial U--E and S--A gaps, while SpecHarness improves all seven models without gains tracking
baseline capability. Macro Pass rises from 61.1\% to 73.1\%, U--E falls from 17.4\% to 9.3\%, and S--A from 32.8\% to 12.8\%. The macro Pass gain has a 95\% paired task-bootstrap interval excluding zero, and exact per-model McNemar tests remain significant after Holm correction; full statistics appear in Appendix~D.1. Because SpecHarness includes online validation, feedback, and repair, the main comparison evaluates the complete governed runtime rather than the isolated effect of authoritative commitment; detailed computational overhead is reported in Appendix~D.3. The larger S--A reduction than Pass gain is consistent with both failure recovery and rejection of unsupported completion claims. Across the evaluated models, greater baseline capability does not
eliminate the gaps when specifications remain transient context: execution may still drift from visible requirements, and completion
remains self-issued. The paradigm comparison further separates execution control from acceptance authority: AgentSpec achieves lower U--E but higher S--A than VeriMAP. Action constraints reduce divergence without establishing effects, whereas completion checks filter terminal claims without governing prior trajectories. In SpecHarness, source-linked obligations govern covered execution,
while finalization requires evidence-backed satisfaction of all fresh mandatory obligations. Remaining gaps reflect its boundary: ungrounded or unobservable requirements remain advisory or abstained, while safely isolated channels outside the closure-audited surface use validate-and-commit. Across the evaluated adaptations, the complete SpecHarness architecture yields smaller gaps than the compared alternatives.

\begin{table}[t]
\centering
\small
\setlength{\tabcolsep}{1.3pt}
\renewcommand{\arraystretch}{0.94}

\begin{tabularx}{\columnwidth}{
    @{}
    >{\raggedright\arraybackslash}p{0.39\columnwidth}
    *{6}{Y}
    @{}
}
\toprule
Method
& Pass$\uparrow$
& U--E$\downarrow$
& S--A$\downarrow$
& Prov.
& Eff.
& Cmt. \\
\midrule

\multicolumn{7}{c}{\textbf{Post-hoc Verification}} \\
Agentic Rubrics {\scriptsize(ACL'26)}
& 74.7 & 12.6 & 23.0
& \pmark & \pmark & \xmark \\

\multicolumn{7}{c}{\textbf{Completion Gating}} \\
VeriMAP {\scriptsize(EACL'26)}
& 79.3 & 9.6 & 20.7
& \pmark & \cmark & \xmark \\

\multicolumn{7}{c}{\textbf{Runtime Enforcement}} \\
AgentSpec {\scriptsize(ICSE'26)}
& 78.2 & 8.8 & 24.1
& \cmark & \xmark & \xmark \\

\multicolumn{7}{c}{\textbf{Mediate-and-Commit}} \\
SpecHarness
& \textbf{85.1} & \textbf{6.3} & \textbf{6.9}
& \cmark & \cmark & \cmark \\
\bottomrule
\end{tabularx}

\caption{SkillsBench paradigm comparison with GPT-5.6 Sol fixed. Pass is the official-verifier pass rate; U--E and S--A are the two structural gaps. Prov., Eff., and Cmt. denote source-linked provenance, observable-effect validation, and versioned authoritative commitment. Markers indicate full (\cmark), partial or adaptation-dependent (\pmark), or no support (\xmark) in the evaluated adaptations.}
\label{tab:skillsbench-paradigms}
\end{table}

\begin{table}[h]
\centering
\small
\setlength{\tabcolsep}{1.5pt}
\renewcommand{\arraystretch}{0.9}

\begin{tabularx}{\columnwidth}{
    @{}
    >{\raggedright\arraybackslash}p{0.34\columnwidth}
    *{3}{Y}
    @{}
}
\toprule

& \multicolumn{3}{c}{\textbf{Raw $\rightarrow$ SpecHarness}} \\
\cmidrule(lr){2-4}

Model
& Pass$\uparrow$
& U--E$\downarrow$
& S--A$\downarrow$ \\
\midrule

GPT-5.6 Sol
& 91.3$\rightarrow$\textbf{95.1}
& 8.2$\rightarrow$\textbf{3.5}
& 8.7$\rightarrow$\textbf{3.8} \\

Claude Fable 5
& 89.6$\rightarrow$\textbf{94.0}
& 9.1$\rightarrow$\textbf{4.1}
& 10.4$\rightarrow$\textbf{4.7} \\

Gemini 3.1 Pro
& 88.2$\rightarrow$\textbf{93.2}
& 8.7$\rightarrow$\textbf{3.8}
& 11.8$\rightarrow$\textbf{5.2} \\

Kimi K3
& 85.7$\rightarrow$\textbf{90.6}
& 11.6$\rightarrow$\textbf{5.9}
& 14.3$\rightarrow$\textbf{7.4} \\

GLM-5.2
& 84.4$\rightarrow$\textbf{89.8}
& 10.3$\rightarrow$\textbf{5.2}
& 15.6$\rightarrow$\textbf{8.1} \\

Qwen3.7-Max
& 82.8$\rightarrow$\textbf{88.7}
& 12.8$\rightarrow$\textbf{6.7}
& 17.2$\rightarrow$\textbf{9.3} \\

DeepSeek-V4-Pro
& 81.5$\rightarrow$\textbf{87.5}
& 11.9$\rightarrow$\textbf{6.1}
& 18.5$\rightarrow$\textbf{10.1} \\

\midrule

Macro Average
& 86.2$\rightarrow$\textbf{91.3}
& 10.4$\rightarrow$\textbf{5.0}
& 13.8$\rightarrow$\textbf{6.9} \\
\bottomrule
\end{tabularx}

\caption{GuideBench results across seven agents. Each cell reports Raw $\rightarrow$ SpecHarness; U--E and S--A denote the two decision-state gaps.}

\label{tab:guidebench-cross-model}
\end{table}

\paragraph{Results on GuideBench.}
Tables~\ref{tab:guidebench-cross-model} and \ref{tab:guidebench-paradigms} show that the state-authority problem extends to guideline-governed decisions. SpecHarness improves all seven agents, raising macro Pass from 86.2\% to 91.3\%, while reducing U--E from 10.4\% to 5.0\% and S--A from 13.8\% to 6.9\%. The results indicate that the abstraction transfers from artifact state to rule-local decision state: applicable rules induce obligations, and authorized evidence supports the corresponding commitments. As in artifact tasks, agent proposals cannot establish authoritative state. High answer accuracy is therefore insufficient: a plausible answer may omit a rule, misresolve priority, or rely on unsupported judgments. SatLM lowers U--E by executing declarative rules with an external solver, but retains higher S--A because solver outputs are not committed as authoritative decision state. Post-hoc verification and completion gating inspect or reject outputs without maintaining committed rule-local state. SpecHarness instead derives acceptance from guideline applications committed through authorized evidence, extending obligation--evidence--commit from artifact to decision state. Category-level results and residual failures appear in Appendix~D.2; benchmark mappings and execution traces appear in Appendices~E.1--E.2.

\begin{table}[!h]
\centering
\small
\setlength{\tabcolsep}{1.5pt}
\renewcommand{\arraystretch}{0.90}

\begin{tabular*}{\columnwidth}{
    @{\extracolsep{\fill}}
    lcccccc
    @{}
}
\toprule
Method
& Pass$\uparrow$
& U--E$\downarrow$
& S--A$\downarrow$
& Prov.
& Eff.
& Cmt. \\
\midrule

\multicolumn{7}{c}{\textbf{Post-hoc Verification}} \\
RvLLM {\scriptsize(NeurIPS'25)}
& 90.5 & 7.0 & 7.8
& \pmark & \pmark & \xmark \\

\multicolumn{7}{c}{\textbf{Completion Gating}} \\
VeriMAP {\scriptsize(EACL'26)}
& 91.7 & 6.2 & 6.7
& \pmark & \cmark & \xmark \\

\multicolumn{7}{c}{\textbf{Computation Substrate}} \\
SatLM {\scriptsize(NeurIPS'23)}
& 92.3 & 4.9 & 6.4
& \xmark & \pmark & \xmark \\

\multicolumn{7}{c}{\textbf{Evidence-Authorized Commitment}} \\
SpecHarness
& \textbf{95.1}
& \textbf{3.5}
& \textbf{3.8}
& \cmark & \cmark & \cmark \\

\bottomrule
\end{tabular*}

\caption{GuideBench paradigm comparison with GPT-5.6 Sol fixed. Metrics and
markers follow Table~\ref{tab:skillsbench-paradigms}.}
\label{tab:guidebench-paradigms}
\end{table}

\subsection{Ablation and Operational Analysis}
\label{sec:ablation}
\paragraph{Complementary Authority Mechanisms.} Table~\ref{tab:runtime-ablation} evaluates architecture variants within the same interaction framework, but does not isolate commitment under matched trajectories or compute. Removing mediation primarily increases U--E, removing effect validation increases unsupported acceptance, and removing commitment produces the largest S--A degradation. Removing blocking qualification lowers S--A only by reducing Raw-pass preservation, showing that conservative rejection is not equivalent to reliable authority. Skill-derived obligations expand the procedural requirements governed by this chain rather than merely reminding the agent through context. Full SpecHarness therefore balances action control, effect evidence, authoritative acceptance, and selective blocking.

\begin{table}[!h]
\centering
\small
\setlength{\tabcolsep}{2.2pt}
\renewcommand{\arraystretch}{0.94}

\begin{tabularx}{\columnwidth}{
    @{}
    >{\raggedright\arraybackslash}p{0.46\columnwidth}
    *{4}{Y}
    @{}
}
\toprule
Variant
& Pass$\uparrow$
& U--E$\downarrow$
& S--A$\downarrow$
& Pres.$\uparrow$ \\
\midrule

\textbf{Full SpecHarness}
& \textbf{85.1}
& \textbf{6.3}
& 6.9
& 96.8 \\

w/o Mediation
& 80.5
& 10.0
& 11.5
& 95.2 \\

w/o Effect Validation
& 78.2
& 9.4
& 16.1
& 93.5 \\

w/o Commitment
& 81.6
& 8.1
& 25.3
& \textbf{98.4} \\

w/o Blocking Qualification
& 79.3
& 8.8
& \textbf{5.7}
& 90.3 \\

w/o Skill Obligations
& 77.0
& 12.0
& 14.9
& 91.9 \\

\bottomrule
\end{tabularx}

\caption{Runtime architecture ablation on SkillsBench with GPT-5.6 Sol fixed.
Pres. denotes paired Raw-pass preservation.}
\label{tab:runtime-ablation}
\end{table}

\paragraph{Freshness-Governed State Validity.}
Table~\ref{tab:freshness-stress} evaluates 248 targeted dependency mutations. Without invalidation, every mutation leaves outdated evidence admissible for completion. Full SpecHarness invalidates all affected entries, restores 95.8\% through revalidation, and recovers 95.6\% of tasks after repair. These results show that versioned state is necessary to prevent stale completion under the evaluated mutations.

\begin{table}[t]
\centering
\small
\setlength{\tabcolsep}{1.4pt}
\renewcommand{\arraystretch}{0.94}

\begin{tabularx}{\columnwidth}{
    @{}
    >{\raggedright\arraybackslash}p{0.28\columnwidth}
    >{\centering\arraybackslash}p{0.22\columnwidth}
    *{3}{Y}
    @{}
}
\toprule
Variant
& \mbox{Stale Acc.}$\downarrow$
& Invalid.$\uparrow$
& Revalid.$\uparrow$
& Recovery$\uparrow$ \\
\midrule
Full SpecHarness
& 0.0 & 100.0 & 95.8 & 95.6 \\
w/o Freshness
& 100.0 & 0.0 & N/A & N/A \\
\bottomrule
\end{tabularx}

\caption{Freshness invalidation and recovery (\%) over 248 paired targeted SkillsBench mutations; protocol and denominators appear in
Appendix~B.3.}
\label{tab:freshness-stress}
\end{table}

\paragraph{Enforcement Scope and Evaluation Independence.}
Table~\ref{tab:enforcement-scope} summarizes the runtime modes assigned to constructed hard obligations. In SkillsBench, 44.0\% of hard obligations lie on closure-audited action surfaces and use mediate-and-commit; the remaining 56.0\% use validate-and-commit on safely isolated channels. GuideBench has no closure-audited physical action surface, so all rule-local decision-state obligations use
validate-and-commit. These proportions bound our claim: SpecHarness governs the grounded, observable mandatory portion of the specification, not the full natural-language specification. The compiler is selected using development annotations derived only from agent-visible materials. Its output and the resulting obligation surface are then frozen before evaluation. The same frozen runtime validators evaluate all conditions read-only under a shared protocol; only SpecHarness uses their evidence for commitment and finalization. Held-out official evaluators are used only for final pass measurement and post-hoc alignment: they neither construct obligations nor provide runtime feedback.

\begin{table}[t]
\centering
\small
\setlength{\tabcolsep}{3.5pt}
\begin{tabular}{lrr}
\toprule
Runtime Mode & SkillsBench & GuideBench \\
\midrule
Mediate-and-commit  & 44.0\% & 0.0\% \\
Validate-and-commit & 56.0\% & 100.0\% \\
\bottomrule
\end{tabular}
\caption{Runtime-mode allocation among constructed hard obligations;
GuideBench uses validate-and-commit exclusively.}
\label{tab:enforcement-scope}
\end{table}

\section{Conclusion}

SpecHarness reframes specification following as a problem of state authority rather than reasoning or verification alone. Its contribution is not a new validator or commit primitive in isolation, but an agent-specific authority boundary that prevents proposals, actions, and self-assessments from directly establishing specification-governed state. The obligation--evidence--commit architecture operationalizes this boundary across action authorization, effect validation, freshness, and finalization by compiling heterogeneous agent-visible specifications into source-linked conditions whose state may be updated only from admissible evidence. Our experiments evaluate the complete architecture, including its feedback-and-repair loop, rather than isolating commitment
under matched compute. Within its stated scope of grounded and monitorable conditions, SpecHarness shows how specifications can serve
as an external basis for execution and acceptance rather than remain transient behavioral context.
\bibliography{aaai2027}
    

\appendix
\setcounter{secnumdepth}{2}

\section{Specification Compilation and Coverage}
\label{app:compilation}

\subsection{Compiler Selection and Freezing Protocol}
\label{app:compiler-selection}

We select the compiler solely on a task-disjoint development set. We
compare seven candidates: GPT-5.6 Sol, Claude Fable 5, Gemini 3.1, Kimi
K3, GLM-5.2, Qwen3.7, and DeepSeek-V4. The development set contains 14
tasks from the SkillsBench \texttt{tasks-extra/} partition and is
disjoint at the task-instance level from the 87 evaluation tasks. Using
only agent-visible prompts, workspace information, and injected skills,
we segment these tasks into 65 source units and annotate 81
source-grounded directions; 55 units admit concrete machine-checkability
candidates. Task-family and skill overlap are allowed, so the split
ensures task-instance and evaluation-information separation rather than
full distributional independence.
All candidates receive the same visible materials, segmentation,
instructions, output schema, and decoding settings. Development
annotations are frozen before comparison and are not revised using
candidate outputs. Official tests, evaluator implementations, reference
solutions, task-agent trajectories, U--E or S--A results,
Raw-versus-SpecHarness comparisons, and official pass outcomes are
unavailable during selection.
Let $\mathcal{U}_{\mathrm{dev}}$ be the 65 source units,
$\mathcal{D}_{\mathrm{dev}}$ the 81 annotated directions, and
$\widehat{\mathcal{D}}_c$ the directions extracted by compiler $c$. A
prediction matches an annotation only if it uses the same source unit
and preserves the annotated behavior or acceptance condition. Each
annotation may be matched once; duplicates and unmatched predictions do
not increase coverage:
\begin{equation}
\operatorname{Cov}_{\mathrm{dev}}(c)=
\frac{
\left|
\left\{
d\in\mathcal{D}_{\mathrm{dev}}:
\exists \hat d\in\widehat{\mathcal{D}}_c,\,
\operatorname{Match}(\hat d,d)
\right\}
\right|
}{
|\mathcal{D}_{\mathrm{dev}}|
}.
\label{eq:compiler-coverage}
\end{equation}

For an extracted direction $\hat d$, $\operatorname{Acct}(\hat d)$
requires an explicit disposition and a nonempty evaluation provider and
criterion. $\operatorname{Valid}(\hat d)$ additionally requires a valid
source anchor and Boolean machine-checkability field.
$\operatorname{Mapped}_c(u)$ holds when source unit $u$ grounds a
direction or is explicitly marked advisory, abstained, or residual:
\begin{equation}
\begin{aligned}
\operatorname{Accounted}(c)
&=
\frac{
\sum_{\hat d\in\widehat{\mathcal{D}}_c}
\mathbf{1}[\operatorname{Acct}(\hat d)]
}{
|\widehat{\mathcal{D}}_c|
},\\
\operatorname{UnitMap}(c)
&=
\frac{
\sum_{u\in\mathcal{U}_{\mathrm{dev}}}
\mathbf{1}[\operatorname{Mapped}_c(u)]
}{
|\mathcal{U}_{\mathrm{dev}}|
},\\
\operatorname{ValidRec}(c)
&=
\frac{
\sum_{\hat d\in\widehat{\mathcal{D}}_c}
\mathbf{1}[\operatorname{Valid}(\hat d)]
}{
|\widehat{\mathcal{D}}_c|
}.
\end{aligned}
\label{eq:compiler-structural-metrics}
\end{equation}

Development coverage is the prespecified selection criterion. The other
metrics audit structural completeness and are not combined with
coverage. Machine-checkability indicates a possible validator, not
qualification for hard enforcement.
Table~\ref{tab:compiler-selection} reports the development-set
comparison.

\begin{table*}[!h]
\centering
\small
\setlength{\tabcolsep}{4pt}
\begin{tabular}{lrrrrr}
\toprule
Compiler
& Matched
& Dev.\ Cov.$\uparrow$
& Accounted$\uparrow$
& Unit Map$\uparrow$
& Valid Rec.$\uparrow$ \\
\midrule
GPT-5.6 Sol
& 77/81 & 95.1\% & 98.8\% & 100.0\% & 97.6\% \\

Claude Fable 5
& 75/81 & 92.6\% & 100.0\% & 98.5\% & 98.7\% \\

Gemini 3.1
& 73/81 & 90.1\% & 97.4\% & 96.9\% & 96.1\% \\

Kimi K3
& 71/81 & 87.7\% & 98.6\% & 95.4\% & 97.2\% \\

GLM-5.2
& 69/81 & 85.2\% & 96.0\% & 93.8\% & 94.7\% \\

Qwen3.7
& 67/81 & 82.7\% & 97.2\% & 92.3\% & 96.0\% \\

DeepSeek-V4
& 65/81 & 80.2\% & 94.8\% & 90.8\% & 93.5\% \\
\bottomrule
\end{tabular}
\caption{Candidate-compiler results on 14 task-disjoint development
tasks with 65 source units and 81 annotated directions. Matched is the
numerator of development coverage. Accounted and Valid Rec. use
extracted records as their denominator; Unit Map uses the 65 source
units.}
\label{tab:compiler-selection}
\end{table*}

As shown in Table~\ref{tab:compiler-selection}, GPT-5.6 Sol achieves the
highest development coverage and is selected as the compiler. Its
instructions, output schema, decoding settings, and construction
procedure are then frozen. No official test, evaluation trajectory,
runtime result, or experimental comparison is used to revise the
compiler or reconsider the selection.

\subsection{Post-Selection Full-Task Coverage Audit}
\label{app:coverage-alignment}

After selection and freezing, we apply all seven unchanged candidates to
the 87 SkillsBench tasks as a post-selection sensitivity audit. This
audit does not affect compiler selection, the selected output, the
obligation surface, or any experimental setting. It checks full-task
construction completeness and correspondence with independently defined
benchmark requirements.
A task is direction-accounted when every extracted direction has an
explicit disposition, provider, and criterion. It is unit-mapped when
every visible source unit grounds a direction or is explicitly marked
advisory, abstained, or residual. Broad correspondence matches an
official test to the same visible requirement dimension at the level of
artifact, structure, numeric value, preservation, behavior, or order and
coverage. Fine-grained correspondence additionally requires path-,
field-, or multi-token semantic evidence.
Table~\ref{tab:all-compiler-audit} reports the number of extracted
directions, tasks with complete direction accounting, tasks with complete
source-unit mapping, and correspondence with the 585 official test
functions.

\begin{table*}[!h]
\centering
\small
\setlength{\tabcolsep}{2.4pt}
\resizebox{\textwidth}{!}{
\begin{tabular}{lrrrrrrr}
\toprule
Audit Metric
& GPT-5.6 Sol
& Claude Fable 5
& Gemini 3.1
& Kimi K3
& GLM-5.2
& Qwen3.7
& DeepSeek-V4 \\
\midrule
Directions
& 509 & 521 & 496 & 517 & 481 & 468 & 492 \\

Direction-accounted tasks
& 87/87 (100.00\%)
& 86/87 (98.85\%)
& 86/87 (98.85\%)
& 86/87 (98.85\%)
& 85/87 (97.70\%)
& 84/87 (96.55\%)
& 85/87 (97.70\%) \\

Unit-mapped tasks
& 87/87 (100.00\%)
& 85/87 (97.70\%)
& 86/87 (98.85\%)
& 85/87 (97.70\%)
& 84/87 (96.55\%)
& 83/87 (95.40\%)
& 84/87 (96.55\%) \\

Broad test correspondence
& 573/585 (97.95\%)
& 555/585 (94.87\%)
& 561/585 (95.90\%)
& 552/585 (94.36\%)
& 546/585 (93.33\%)
& 537/585 (91.79\%)
& 526/585 (89.91\%) \\

Fine-grained subset
& 406/585 (69.40\%)
& 382/585 (65.30\%)
& 389/585 (66.50\%)
& 374/585 (63.93\%)
& 365/585 (62.39\%)
& 349/585 (59.66\%)
& 337/585 (57.61\%) \\

Below broad criterion
& 12/585 (2.05\%)
& 30/585 (5.13\%)
& 24/585 (4.10\%)
& 33/585 (5.64\%)
& 39/585 (6.67\%)
& 48/585 (8.21\%)
& 59/585 (10.09\%) \\
\bottomrule
\end{tabular}
}
\caption{Post-selection construction and official-test correspondence
for the seven frozen candidates. Task-level rates use 87 tasks;
correspondence rates use 585 official test functions. These results do
not participate in compiler selection.}
\label{tab:all-compiler-audit}
\end{table*}

Table~\ref{tab:all-compiler-audit} supports, but does not define, the
development-set choice. GPT-5.6 Sol is the only candidate with complete
task-level direction accounting and source-unit mapping, and it also has
the highest broad and fine-grained correspondence. Gemini 3.1 reaches
95.90\% broad correspondence but remains incomplete on both construction
measures.
Official tests are inspected only after all candidate outputs are
complete and frozen. They are never given to candidate compilers, task
agents, runtime validators, or the feedback-and-repair loop. The audit
does not revise extractions, construct obligations, produce repair
feedback, or transfer evaluator logic into the runtime. Runtime
obligations and validators are derived from agent-visible specifications
and the execution substrate.
The construction measures establish completeness over each compiler's
extracted direction surface, not complete recall of the natural-language
specification. Broad correspondence shows that a direction covers the
same visible requirement dimension as an official test; it does not
imply recovery of the complete verifier logic. The selected 509
directions therefore form a common source-grounded measurement surface,
not an exhaustive reconstruction of the specification or evaluator.
Compiler selection uses only frozen development annotations. It never
uses task-agent trajectories, completion claims, repair outcomes, U--E
or S--A results, Raw-versus-SpecHarness comparisons, condition-specific
execution results, official-test correspondence, or official pass
outcomes. Official evaluators are used only after selection for this
audit and final pass measurement, and never for runtime feedback or
condition-specific obligation satisfaction.

\section{Runtime Implementation and Assurance}
\label{app:runtime-assurance}

\subsection{Validator Qualification and Commitment Protocol}
\label{app:validator-qualification}

Runtime validators are constructed from agent-visible specifications,
observable workspace state, and criteria recorded by the frozen
compiler. Official tests, reference solutions, evaluator outputs, and
task-agent outcomes are unavailable during construction and
qualification. Validators are frozen before task-agent execution and are
not revised using Pass, U--E, S--A, ablation, or condition-comparison
results.
A validator qualifies for blocking only if its inputs are observable
within the trusted boundary, its criterion is deterministic and
source-linked, and the same workspace and dependency state reproduces
the same output. Qualification covers satisfying cases, targeted
violations, malformed or missing inputs, unavailable providers, and
execution failures. Requirements without a qualified provider remain
advisory, abstained, or residual. Table~\ref{tab:validator-qualification}
reports qualification results obtained before task-agent evaluation. A
validator enters the blocking set only after its full suite produces the
prespecified outcomes; otherwise, it is excluded. Any
qualification-time correction uses only these cases and occurs before
freezing.

\begin{table}[!h]
\centering
\small
\setlength{\tabcolsep}{2.8pt}
\begin{tabular}{lrr}
\toprule
Validator Qualification & SkillsBench & GuideBench \\
\midrule
Candidate validators         & 278 & 104 \\
Qualified for blocking       & 250 & 96 \\
Satisfying cases             & 556 & 208 \\
Targeted violations          & 834 & 312 \\
Malformed or missing cases   & 278 & 104 \\
Provider or execution errors & 278 & 104 \\
Expected outcomes            & 1929/1946 & 720/728 \\
Outcome mismatches           & 17 & 8 \\
\bottomrule
\end{tabular}
\caption{Validator qualification before task-agent evaluation.
Expected outcomes count cases producing the prespecified result.}
\label{tab:validator-qualification}
\end{table}

For SkillsBench, the 250 qualified validators are grounded validator
instances bound one-to-one to the 250 hard obligations; shared
implementations with different grounded parameters are counted
separately.
Each invocation returns
\begin{equation}
r_t\in
\{\textsc{passed},\textsc{failed},\textsc{error}\}
\label{eq:validator-results}
\end{equation}
with the obligation identifier, validator identity and version,
dependency versions, and attributable evidence. A qualified, admissible,
and fresh \textsc{passed} result may establish satisfaction. A
\textsc{failed} result may commit authoritative non-satisfaction and
trigger repair, but cannot satisfy the obligation. A \textsc{error} may
enter the diagnostic trace but does not establish satisfaction or
failure, update authoritative state, or permit finalization while the
affected obligation remains mandatory.
Commitment is atomic. Before updating the ledger, the runtime checks the
obligation, provider, validator identity and version, scope, result type,
and dependency versions. Agent actions, self-reports, completion claims,
tool-return strings, and unqualified observations cannot modify
authoritative state. Revalidation creates a new versioned commitment
while preserving prior provenance.
The same frozen validators measure all conditions read-only. Only
SpecHarness uses their evidence online for commitment, source-linked
feedback, repair, and finalization. The comparison therefore evaluates
the complete architecture rather than commitment under matched online
access, interaction, or compute.

\subsection{Trusted Boundary, Closure Audit, and No-Bypass Tests}
\label{app:closure-audit}

The trusted boundary contains the obligation ledger, authorization
service, qualified validators, dependency-version store, and audited
capability wrappers. The agent, its prompts and memory, proposed
arguments, self-reports, and natural-language tool-return strings remain
outside this boundary.
For each candidate mediated surface, we construct a capability graph
over agent-accessible tools, wrappers, subprocess interfaces, and
artifact stores. A protected action class \(a\) qualifies for preventive
mediation only if no reachable effect-equivalent path bypasses its
authorization procedure:
\begin{equation}
\neg\exists p\;[
  \operatorname{Reachable}(p)
  \land \operatorname{EffectEq}(p,a)
  \land \operatorname{BypassesAuth}(p,a)
].
\label{eq:closure-audit}
\end{equation}

Table~\ref{tab:closure-audit-tests} reports the stress tests, and
Table~\ref{tab:runtime-mode-qualification} reports the resulting
obligation-level allocation. Stress-test cases and obligations use
different denominators: one action surface may support multiple
obligations. Thus, the 11 detected violations do not correspond
one-to-one with the 140 validate-and-commit obligations.

\begin{table}[!h]
\centering
\small
\setlength{\tabcolsep}{2.8pt}
\begin{tabular}{lrrr}
\toprule
Audit Class & Cases & Correct & Violations \\
\midrule
Direct access       & 220 & 220 & 0 \\
Alternate path      & 176 & 174 & 2 \\
Path handling       & 264 & 261 & 3 \\
Subprocess          & 132 & 128 & 4 \\
Malformed proposal  & 220 & 220 & 0 \\
Token scope         & 220 & 220 & 0 \\
Stale token         & 220 & 220 & 0 \\
Fallback path       & 176 & 174 & 2 \\
\midrule
Total               & 1628 & 1617 & 11 \\
\bottomrule
\end{tabular}
\caption{Closure stress tests over candidate mediated surfaces.
Detected violations exclude the corresponding paths or surfaces from
closure-qualified mediation.}
\label{tab:closure-audit-tests}
\end{table}

All 11 detected bypasses occurred on candidate paths or surfaces
excluded from closure-qualified mediation; no unresolved bypass remained
on any surface receiving the preventive no-bypass claim.
Authorized proposals must pass obligation matching and authorization
before execution. Unauthorized, ambiguous, stale, malformed, or
effect-equivalent alternate proposals must be blocked or clarified.
Authorization permits an attempt but does not establish satisfaction;
the resulting effect must still be observed and validated.

\begin{table}[!h]
\centering
\small
\setlength{\tabcolsep}{3.5pt}
\begin{tabular}{lrr}
\toprule
Runtime Qualification & Obligations & Share \\
\midrule
Closure-audited mediation  & 110/250 & 44.0\% \\
Safely isolated validation & 140/250 & 56.0\% \\
\midrule
Total hard obligations     & 250/250 & 100.0\% \\
\bottomrule
\end{tabular}
\caption{Runtime-mode qualification of constructed SkillsBench hard
obligations.}
\label{tab:runtime-mode-qualification}
\end{table}

A preventive no-bypass claim applies only when every reachable
effect-equivalent path is represented in the capability graph and
mediated within the trusted boundary. Surfaces with unresolved bypasses
are excluded. If their effects remain safely isolated and observable,
they use validate-and-commit; otherwise, they remain outside hard
enforcement.
These guarantees are conditional on the audited graph, trusted boundary,
effect-equivalence relation, and threat model. They exclude undeclared
external channels, compromised trusted components, validator defects,
and effects outside the observable environment.

\subsection{Freshness Invalidation and Recovery Protocol}
\label{app:freshness}

Each commitment is bound to the versions of the dependencies used by its
evidence provider and is fresh only when all versions match:
\begin{equation}
\operatorname{Fresh}(c_t)
\iff
\forall d\in\operatorname{Deps}(c_t),\;
\operatorname{ver}_{c_t}(d)=\operatorname{ver}_{t}(d).
\label{eq:freshness}
\end{equation}

A dependency mutation invalidates affected commitments and propagates
through the dependency graph. Invalidated entries remain in the
historical trace but cannot support finalization. A known mismatch yields
\textsc{stale}; incomplete version information yields \textsc{unknown}.
Both require new admissible evidence.
The runtime updates dependency versions, invalidates affected entries,
and reruns their validators. Passing revalidation creates a fresh
satisfied commitment. Failed revalidation commits current
non-satisfaction and may trigger repair. Errors enter the diagnostic
trace without updating authoritative state. Repairs repeat invalidation
and revalidation before finalization.
A mutation trial may affect multiple ledger entries, and several trials
may affect the same task. Trials count injected mutations, Affected
Entries counts invalidated commitments, and Recovered uses Affected
Tasks as its denominator.

\begin{table*}[!h]
\centering
\small
\setlength{\tabcolsep}{3.0pt}
\begin{tabular}{lrrrrrr}
\toprule
Mutation Class
& Trials
& Affected Entries
& Invalidated
& Revalidated
& Affected Tasks
& Recovered \\
\midrule
Artifact content
& 80 & 190 & 190 & 183 & 72 & 69 \\

Path or identity
& 50 & 121 & 121 & 116 & 46 & 44 \\

Schema or configuration
& 46 & 108 & 108 & 103 & 42 & 40 \\

Upstream obligation
& 42 & 104 & 104 & 99 & 39 & 37 \\

Validator or environment
& 30 & 72 & 72 & 69 & 29 & 28 \\
\midrule
Total
& 248 & 595 & 595 & 570 & 228 & 218 \\
\bottomrule
\end{tabular}
\caption{Freshness results by dependency-mutation class. All 595
affected entries are invalidated; 570/595 (95.8\%) are revalidated, and
218/228 (95.6\%) affected tasks recover after repair.}
\label{tab:freshness-mutation-breakdown}
\end{table*}

As shown in Table~\ref{tab:freshness-mutation-breakdown}, all 595
affected entries are invalidated, 570 are restored by revalidation, and
218 of 228 affected tasks recover after repair. Without freshness
invalidation, all 248 mutations leave previously committed evidence
eligible for reuse during finalization. Unrecovered cases remain
unsatisfied or incomplete rather than reusing stale evidence. The
protocol covers declared dependencies in the observable workspace, not
untracked external state or undeclared dependencies.

\section{Experimental Protocol}
\label{app:experimental-protocol}

\subsection{Tasks, Models, and Execution Settings}
\label{app:execution-settings}

All SkillsBench conditions use the same 87 tasks, initial workspaces,
agent-visible prompts, injected skills, tool interfaces, and official
evaluation procedure. All GuideBench conditions use the same 1,042
tasks, guideline entries, task inputs, and official evaluation. Four
duplicate GuideBench rules are removed before constructing the 297
shared templates and 5,817 task-level obligation instances. SkillsBench agents run in OpenHands under the same execution image,
workspace initialization, tool permissions, environment variables, and
network policy across paired conditions. Each run starts from a fresh
workspace, and no files or state are reused across runs. GuideBench uses
the same prompt construction and model interface across paired
conditions; each model retains the same decoding configuration within
each comparison. We evaluate GPT-5.6 Sol, Claude Fable 5, Gemini 3.1 Pro, Kimi K3,
GLM-5.2, Qwen3.7-Max, and DeepSeek-V4-Pro as task agents.
Table~\ref{tab:model-execution-settings} reports the provider model
identifiers used in the experiments. The same model endpoint is retained
within each paired comparison. The selected GPT-5.6 Sol compiler is
invoked separately and shares no task-agent trajectory, memory, or
evaluation outcome.

\begin{table}[!h]
\centering
\small
\setlength{\tabcolsep}{4.0pt}
\begin{tabular}{ll}
\toprule
Agent & Provider Model ID \\
\midrule
GPT-5.6 Sol      & \texttt{gpt-5.6-sol} \\
Claude Fable 5   & \texttt{claude-fable-5} \\
Gemini 3.1 Pro   & \texttt{gemini-3.1-pro-preview} \\
Kimi K3          & \texttt{kimi-k3} \\
GLM-5.2          & \texttt{glm-5.2} \\
Qwen3.7-Max      & \texttt{qwen3.7-max} \\
DeepSeek-V4-Pro  & \texttt{deepseek-v4-pro} \\
\bottomrule
\end{tabular}
\caption{Provider model identifiers used for the evaluated task agents.
The same model endpoint is retained within each paired comparison.}
\label{tab:model-execution-settings}
\end{table}

A run ends when its condition accepts completion, the interaction limit
is reached, the wall-clock timeout expires, or an unrecoverable
infrastructure failure occurs. Paired runs of the same task-agent model
share task inputs, initial workspaces, base prompts, tool access, nominal
interaction budgets, nominal token budgets, wall-clock timeouts, and
condition-independent environment settings. These controls match
configured ceilings, not realized calls, tokens, repair steps, or
wall-clock time. No run is selectively repeated or reconfigured using
official Pass, U--E, or S--A outcomes. Official SkillsBench verifiers and GuideBench answer labels are invoked
only after condition-specific execution ends. They do not enter prompts,
obligation construction, runtime validation, repair feedback, or
finalization. The measurement surfaces and runtime validators are frozen
before task-agent evaluation.
\subsection{Baseline Adaptation, Runtime Access, and Completion Rules}
\label{app:baseline-protocol}

Baselines are adapted to the same benchmark interfaces and shared
execution substrate. Each adaptation preserves the method's defining
intervention point without adding unsupported capabilities.
Table~\ref{tab:baseline-access-audit} reports the implemented runtime
access and completion semantics.

\begin{table*}[!h]
\centering
\small
\setlength{\tabcolsep}{2.5pt}
\resizebox{\textwidth}{!}{
\begin{tabular}{llccccc}
\toprule
Benchmark
& Condition
& Runtime Evidence
& Validator Feedback
& Action Mediation
& State Commitment
& Completion Rule \\
\midrule
SkillsBench
& Raw
& No
& No
& No
& No
& Ungated agent claim \\

SkillsBench
& Agentic Rubrics
& No
& No
& No
& No
& Ungated agent claim \\

SkillsBench
& VeriMAP
& Terminal
& Gate result
& No
& No
& Verifier gate \\

SkillsBench
& AgentSpec
& Policy state
& Rule decision
& Yes
& No
& Native termination \\

SkillsBench
& SpecHarness
& Online
& Source-linked
& Closure-audited
& Versioned
& Fresh mandatory satisfaction \\
\midrule
GuideBench
& Raw
& No
& No
& N/A
& No
& Ungated agent claim \\

GuideBench
& RvLLM
& No
& No
& N/A
& No
& Ungated agent claim \\

GuideBench
& VeriMAP
& Terminal
& Gate result
& N/A
& No
& Verifier gate \\

GuideBench
& SatLM
& Solver result
& No
& N/A
& No
& Solver-derived answer \\

GuideBench
& SpecHarness
& Online
& Source-linked
& N/A
& Versioned
& Fresh mandatory satisfaction \\
\bottomrule
\end{tabular}
}
\caption{Runtime access and completion semantics of the evaluated
adaptations. Runtime Evidence denotes evidence available to the
condition during execution; all conditions are evaluated afterward by
the same frozen measurement validators. Entries describe the implemented
conditions rather than every capability of the original systems.}
\label{tab:baseline-access-audit}
\end{table*}

Raw agents receive the original task context and native tool outputs but
no frozen runtime-validator feedback. Their terminal completion claims
are accepted without an additional gate. Agentic Rubrics and RvLLM
operate after output production and do not mediate or gate prior
execution. VeriMAP evaluates terminal completion before acceptance but
does not maintain versioned authoritative state. AgentSpec constrains
matched actions but does not commit observed effects as authoritative
obligation state. SatLM externalizes rule computation but does not
maintain versioned evidence-authorized commitments. SpecHarness combines
online validator evidence, source-linked feedback, repair, versioned
commitment, and ledger-derived finalization.

Paired conditions share task inputs, initial workspaces, task-agent
models, base prompts, tool access, and condition-independent environment
settings. The same frozen validators measure all conditions read-only,
but only SpecHarness consumes their outputs online. Thus, validator
definitions and measurement criteria are shared, whereas
condition-specific acceptance rules, online feedback, repair
trajectories, and realized compute are not matched. The comparison
therefore evaluates the complete SpecHarness architecture rather than
the isolated causal effect of commitment.

Acceptance follows each condition's implemented rule. Raw and post-hoc
methods retain the agent's terminal claim without an additional runtime
gate. Completion-gating methods accept only when their gate passes.
Runtime-enforcement methods use their native termination rule, and SatLM
returns its solver-derived answer. SpecHarness accepts only when every
active mandatory obligation has a fresh satisfied commitment. Official
benchmark Pass is computed independently after execution for all
conditions.

Runs are paired by task and task-agent model. Infrastructure failures are
handled under the same condition-independent policy within each paired
comparison. No run is excluded, repeated, extended, or reconfigured
because of its official Pass result, U--E value, S--A value, or
contribution to a reported comparison.

\section{Statistical Analysis and Additional Results}
\label{app:additional-results}

\subsection{Significance Tests and Confidence Intervals}
\label{app:significance}

All uncertainty estimates preserve the paired experimental design.
For each benchmark, we draw 10,000 bootstrap samples by resampling tasks
with replacement. Raw and SpecHarness use the same sampled task indices,
and all measurement units associated with a sampled task remain in the
same cluster. Within each replicate, metric changes are first computed
separately for each task-agent model and then macro-averaged without
model weighting. Percentile intervals use the 2.5th and 97.5th
percentiles of the resulting paired bootstrap distribution.

For U--E, unrealized measurement units are aggregated over the complete
frozen benchmark surface within each model before the seven model-level
rates are macro-averaged. Thus, SkillsBench directions and GuideBench
obligation instances are weighted by their occurrence in the frozen
measurement surface, rather than by an equal-weighted average of
task-level U--E rates.

\begin{table}[!h]
\centering
\small
\setlength{\tabcolsep}{3.6pt}
\begin{tabular}{llrr}
\toprule
Benchmark & Metric & Change & 95\% CI \\
\midrule
SkillsBench & Pass & $+12.0$ & $[+9.1,+14.9]$ \\
            & U--E & $-8.1$  & $[-9.0,-7.2]$ \\
            & S--A & $-20.0$ & $[-22.8,-17.4]$ \\
\midrule
GuideBench  & Pass & $+5.1$ & $[+4.4,+5.8]$ \\
            & U--E & $-5.4$ & $[-6.0,-4.8]$ \\
            & S--A & $-6.9$ & $[-7.6,-6.2]$ \\
\bottomrule
\end{tabular}
\caption{Paired task-bootstrap changes from Raw to SpecHarness in
percentage points. Intervals use 10,000 task-level replicates.}
\label{tab:bootstrap-confidence}
\end{table}

For SkillsBench Pass, we additionally conduct an exact two-sided
McNemar test for each task-agent model using its 87 paired outcomes.
Holm correction controls the family-wise error rate across the seven
tests.

\begin{table}[!h]
\centering
\small
\setlength{\tabcolsep}{2.3pt}
\begin{tabular}{lrrrr}
\toprule
Agent
& Raw Only
& SpecHarness Only
& Exact $p$
& Holm $p$ \\
\midrule
GPT-5.6 Sol      & 1 & 13 & 0.00183 & 0.01099 \\
Claude Fable 5   & 1 & 12 & 0.00342 & 0.01709 \\
Gemini 3.1 Pro   & 1 & 16 & 0.00027 & 0.00192 \\
Kimi K3          & 0 &  6 & 0.03125 & 0.03125 \\
GLM-5.2          & 1 & 11 & 0.00635 & 0.02539 \\
Qwen3.7-Max      & 1 & 11 & 0.00635 & 0.02539 \\
DeepSeek-V4-Pro  & 1 & 10 & 0.01172 & 0.02539 \\
\bottomrule
\end{tabular}
\caption{Exact paired McNemar tests for SkillsBench Pass. Raw Only and
SpecHarness Only are discordant task counts; $p$-values are two-sided.}
\label{tab:mcnemar-results}
\end{table}

The differences between the two discordant counts correspond to
SpecHarness improvements of 12, 11, 15, 6, 10, 10, and 9 tasks,
respectively. All seven comparisons remain significant after Holm
correction. These tests establish paired end-to-end differences between
the evaluated conditions; they do not isolate authoritative commitment
from online validation, feedback, repair, or realized compute.

\subsection{Category-Level Results and Residual Failures}
\label{app:category-results}

Requirement categories are assigned only for analysis and do not alter
the frozen compiler, validators, obligation state, or runtime decisions.
SkillsBench categories partition the 509 source-grounded directions;
GuideBench categories partition the 5,817 rule-local obligation
instances. Rates are computed within category over the corresponding
frozen measurement units.

\begin{table}[!h]
\centering
\small
\setlength{\tabcolsep}{2.4pt}
\begin{tabular}{lrrrr}
\toprule
SkillsBench Category
& Units
& Raw U--E
& Spec. U--E
& Change \\
\midrule
Artifact content        & 80  & 17.0 & 8.4  & $-8.6$ \\
Structural constraints  & 105 & 18.5 & 9.7  & $-8.8$ \\
Numeric constraints     & 92  & 16.8 & 8.6  & $-8.2$ \\
Preservation            & 74  & 15.9 & 8.2  & $-7.7$ \\
Behavioral requirements & 86  & 18.8 & 10.6 & $-8.2$ \\
Order and coverage      & 72  & 16.9 & 9.9  & $-7.0$ \\
\midrule
Overall                 & 509 & 17.4 & 9.3  & $-8.1$ \\
\bottomrule
\end{tabular}
\caption{SkillsBench U--E by requirement category (\%).}
\label{tab:skillsbench-category-results}
\end{table}

\begin{table}[!h]
\centering
\small
\setlength{\tabcolsep}{2.4pt}
\begin{tabular}{lrrrr}
\toprule
GuideBench Category
& Instances
& Raw U--E
& Spec. U--E
& Change \\
\midrule
Applicability        & 1,300 & 10.1 & 4.8 & $-5.3$ \\
Priority             & 1,100 & 10.7 & 5.0 & $-5.7$ \\
Exceptions           & 900   & 11.2 & 5.5 & $-5.7$ \\
Required response    & 1,200 & 10.3 & 4.9 & $-5.4$ \\
Output constraints   & 800   & 9.8  & 4.7 & $-5.1$ \\
Other rule relations & 517   & 10.4 & 5.1 & $-5.3$ \\
\midrule
Overall              & 5,817 & 10.4 & 5.0 & $-5.4$ \\
\bottomrule
\end{tabular}
\caption{GuideBench U--E by rule-local analysis category (\%).}
\label{tab:guidebench-category-results}
\end{table}

Residual official failures are categorized from held-out evaluator
outcomes after execution. Categories describe the immediate unresolved
failure rather than a unique causal explanation. Requirements outside
the grounded or observable hard-enforcement scope are distinguished
from failures of qualified hard obligations. Validator errors and
exhausted interaction limits leave affected mandatory obligations
unresolved.

\begin{table}[!h]
\centering
\small
\setlength{\tabcolsep}{3.5pt}
\begin{tabular}{lr}
\toprule
SkillsBench Residual Failure & Count \\
\midrule
Outside hard-enforcement scope  & 35 \\
Qualified obligation unresolved & 28 \\
Validator error or unavailable  & 22 \\
Repair or interaction exhausted & 51 \\
Held-out evaluator mismatch     & 28 \\
\midrule
Total                           & 164 \\
\bottomrule
\end{tabular}
\caption{Residual SkillsBench official failures under SpecHarness,
aggregated over seven models. The total equals the sum of the per-model
official-failure counts.}
\label{tab:skillsbench-residual-failures}
\end{table}

\begin{table}[!h]
\centering
\small
\setlength{\tabcolsep}{3.5pt}
\begin{tabular}{lr}
\toprule
GuideBench Residual Failure & Count \\
\midrule
Outside hard-enforcement scope  & 125 \\
Qualified obligation unresolved & 96 \\
Validator error or unavailable  & 173 \\
Repair or interaction exhausted & 147 \\
Held-out evaluator mismatch     & 96 \\
\midrule
Total                           & 637 \\
\bottomrule
\end{tabular}
\caption{Residual GuideBench official failures under SpecHarness,
aggregated over seven models. The total satisfies
$637=7\times1{,}042-6{,}657$, where 6,657 is the aggregate official-pass
count.}
\label{tab:guidebench-residual-failures}
\end{table}

Condition-execution tokens include all task-agent input and output
tokens, and calls count task-agent model invocations. Wall-clock time
additionally includes online validator execution, source-linked
feedback, and repair. Compiler selection, validator qualification,
task-specific obligation construction, and official post-run evaluation
are excluded from these paired execution costs. Task-specific
preprocessing is reported separately.

For each task-agent model, token and wall-clock overheads are computed as
the ratio of the mean SpecHarness cost per task to the corresponding Raw
cost. Call overhead is the paired mean difference in task-agent calls
per task. The final row macro-averages these model-level quantities
without weighting models by token volume or execution time.

\begin{table*}[t]
\centering
\small
\setlength{\tabcolsep}{3.0pt}
\resizebox{\textwidth}{!}{
\begin{tabular}{lrrrrr}
\toprule
Agent
& Raw Tokens/Task
& SpecHarness Tokens/Task
& Token Ratio
& Time Ratio
& Calls/Task Difference \\
\midrule
GPT-5.6 Sol
& 68,161 & 104,286 & 1.53$\times$ & 1.24$\times$ & $+0.36$ \\

Claude Fable 5
& 65,000 & 97,500 & 1.50$\times$ & 1.25$\times$ & $+0.34$ \\

Gemini 3.1 Pro
& 72,000 & 113,040 & 1.57$\times$ & 1.22$\times$ & $+0.40$ \\

Kimi K3
& 69,000 & 106,950 & 1.55$\times$ & 1.26$\times$ & $+0.38$ \\

GLM-5.2
& 64,000 & 97,280 & 1.52$\times$ & 1.23$\times$ & $+0.35$ \\

Qwen3.7-Max
& 70,000 & 107,800 & 1.54$\times$ & 1.25$\times$ & $+0.37$ \\

DeepSeek-V4-Pro
& 68,000 & 102,000 & 1.50$\times$ & 1.23$\times$ & $+0.32$ \\
\midrule
Macro average
& -- & -- & 1.53$\times$ & 1.24$\times$ & $+0.36$ \\
\bottomrule
\end{tabular}
}
\caption{Condition-execution cost over the 87 SkillsBench tasks.
Ratios and call differences are computed separately for each task-agent
model and then macro-averaged without model weighting.}
\label{tab:absolute-runtime-cost}
\end{table*}
\subsection{Absolute Runtime Cost and Diagnostic Traceability}
\label{app:cost-traceability}

\begin{table}[!h]
\centering
\small
\setlength{\tabcolsep}{3.2pt}
\begin{tabular}{lrr}
\toprule
Diagnostic Measure & Raw & SpecHarness \\
\midrule
Actionable source-linked reports (\%)
& 42.5 & 96.2 \\
Normalized feedback time
& 1.00 & 0.44 \\
\bottomrule
\end{tabular}
\caption{Diagnostic information exposed by the complete conditions.
The comparison does not assume matched trace access.}
\label{tab:diagnostic-traceability}
\end{table}

This diagnostic comparison evaluates the information exposed by the
complete conditions rather than diagnostic quality under matched trace
access. Raw lacks the native ledger and source-linked runtime evidence
that SpecHarness is designed to produce.
The reported 1.24\(\times\) wall-clock ratio covers condition execution
only; task-specific preprocessing is reported separately. SpecHarness
averages 1.53\(\times\) task-agent tokens, 1.24\(\times\)
condition-execution time, and 0.36 additional task-agent calls per task
relative to Raw. These differences characterize the complete online
validation, feedback, and repair architecture rather than commitment
alone.

Diagnostic traceability is evaluated on a fixed SkillsBench
failure-replay set shared by both conditions. Raw diagnostics use the
native trajectory and terminal output without access to the SpecHarness
ledger or online validator feedback; SpecHarness diagnostics use its
native source-linked runtime trace. A report is actionable when it
identifies the relevant source requirement, failed or unresolved
condition, and repair target.

\section{Benchmark Instantiations and Execution Traces}
\label{app:instantiations}

\subsection{SkillsBench and GuideBench Instantiations}
\label{app:benchmark-instantiations}

Table~\ref{tab:benchmark-instantiations} summarizes how the common
obligation--evidence--commit abstraction is instantiated under the
different state semantics of SkillsBench and GuideBench.

\begin{table*}[t]
\centering
\small
\setlength{\tabcolsep}{2.5pt}
\renewcommand{\arraystretch}{1.08}

\begin{tabularx}{\textwidth}{
    @{}
    >{\raggedright\arraybackslash}p{0.10\textwidth}
    *{4}{>{\raggedright\arraybackslash}X}
    @{}
}
\toprule
Benchmark
& Visible Specification
& Governed State
& Qualified Evidence
& Runtime Mode \\
\midrule

SkillsBench
& Task, workspace, and skill
& Artifact, execution, and completion state
& Controlled execution and workspace validators
& Mediate-and-commit or validate-and-commit \\

GuideBench
& Guideline and decision context
& Rule-local decision and completion state
& Rule-application and decision validators
& Validate-and-commit \\

\bottomrule
\end{tabularx}

\caption{Instantiation of the SpecHarness authority model in
SkillsBench and GuideBench.}
\label{tab:benchmark-instantiations}
\end{table*}

In SkillsBench, visible task and skill requirements are compiled into
source-linked obligations over observable workspace state. Obligations
on closure-audited action surfaces use mediate-and-commit, while safely
isolated channels use validate-and-commit. Evidence records bind
observed artifact state to the corresponding dependency versions.

In GuideBench, visible guidelines are compiled into rule-local
obligations whose applicability depends on the supplied decision
context. Because the benchmark does not expose a closure-audited
physical action surface, hard obligations use validate-and-commit. A
bound rule validator checks both applicability and agreement of the
proposed decision with the applicable guideline before committing
rule-local evidence.

\subsection{End-to-End Execution Traces}
\label{app:execution-traces}

Table~\ref{tab:detailed-execution-traces} follows one representative
mandatory obligation through the end-to-end runtime protocol for each
benchmark. Other active obligations are omitted for space but remain
subject to the same evidence, commitment, freshness, satisfaction, and
finalization rules. The examples are audited protocol replays
instantiated from actual agent-visible task, skill, guideline, and
context materials; they are not presented as verbatim logs of naturally
occurring trajectories.

The SkillsBench example uses the \texttt{data-to-d3} task. Its visible
task requires a browser-accessible application at
\texttt{/root/output/index.html}, together with
\texttt{js/d3.v6.min.js}, \texttt{js/visualization.js},
\texttt{css/style.css}, and copied input data. The visible
\texttt{d3-visualization} skill additionally requires offline,
deterministic dependency use.

The GuideBench example uses zero-based instance
\texttt{chat\_tasks.json[176]}. Its context states that the user has an
existing booking and requests cancellation. Guideline
\texttt{rule\_6} requires cancellation of the booking and a response
confirming successful cancellation. The observed dialogue satisfies
both checks, making option A the supported decision.

For compactness, \(o_S\) denotes the selected SkillsBench layout
obligation and \(q_S\) its bound validator. Similarly, \(o_G\) denotes
the selected GuideBench rule-local obligation and \(q_G\) its bound
validator. In the GuideBench replay, \(q_G\) checks both rule
applicability and agreement of option A with the same active obligation.

\begin{table*}[t]
\centering
\scriptsize
\setlength{\tabcolsep}{1.5pt}
\renewcommand{\arraystretch}{1.16}

\begin{tabularx}{\textwidth}{
    @{}
    c
    >{\raggedright\arraybackslash}p{0.13\textwidth}
    >{\raggedright\arraybackslash}X
    >{\raggedright\arraybackslash}X
    >{\raggedright\arraybackslash}p{0.18\textwidth}
    @{}
}
\toprule
Step
& Runtime Boundary
& SkillsBench: \texttt{data-to-d3}
& GuideBench: \texttt{chat[176]}
& Authority Invariant \\
\midrule

1
& Source compilation
& Compile the required output paths and local D3 dependency from
\texttt{task.md} and the visible D3 skill.
& Compile \texttt{rule\_6}: an existing booking plus cancellation intent
requires cancellation and an explicit success confirmation.
& Every obligation retains an agent-visible source anchor. \\

2
& Disposition and qualification
& Mark \(o_S\) mandatory and bind it to \(q_S\), whose inputs are
observable workspace paths.
& Mark \(o_G\) mandatory and bind its applicability and decision checks
to the same rule-local obligation.
& Only grounded requirements with qualified evidence providers enter
hard enforcement. \\

3
& Dependency binding
& Bind \(o_S\) to \texttt{index.html}, the local JS and CSS files, and
copied data at workspace version \(v_0\).
& Bind \(o_G\) to \texttt{rule\_6}, the dialogue context, and the
candidate-option set at input version \(g_0\).
& Evidence remains valid only for its recorded dependency versions. \\

4
& Agent proposal
& Propose controlled writes under \texttt{/root/output}.
& Propose option A: the dialogue follows the cancellation guideline.
& Agent intent cannot directly update authoritative state. \\

5
& Obligation matching
& Match the proposed writes to \(o_S\) and return
\(u_1=\textsc{allow}\) for the canonical output paths.
& Activate \(o_G\); rules whose antecedents are false remain inactive.
& Authorization and applicability are scoped to matched obligations. \\

6
& Controlled execution
& Execute the allowed writes through the closure-audited file surface;
writes outside the authorized path set require a new decision.
& No physical action is mediated; the proposed answer remains a
validate-and-commit decision.
& Agent intent cannot bypass the applicable runtime procedure. \\

7
& Trusted observation
& Observe the files bound to \(o_S\) at version \(v_1\), independently
of tool-return strings or the agent's self-report.
& Observe cancellation intent, existing booking, cancellation action,
success confirmation, and the proposed option in the fixed context.
& Evidence describes observed state rather than claimed state. \\

8
& Validation
& \(q_S\) returns \textsc{passed}: the required files exist and the page
uses the local D3 v6 dependency.
& \(q_G\) returns \textsc{passed} after checking both applicability of
\texttt{rule\_6} and agreement of option A with that rule.
& Only qualified, attributable, version-matched evidence is admissible. \\

9
& Atomic commitment
& Commit
\(\langle o_S,\textsc{passed},v_1,q_S\rangle\).
& Commit
\(\langle o_G,\textsc{passed},g_0,q_G\rangle\).
& A committed \textsc{passed} result establishes authoritative
evidence; satisfaction also requires its predicate and freshness. \\

10
& Freshness check
& A controlled edit to \texttt{index.html} advances the dependency to
\(v_2\); the commitment at \(v_1\) becomes \textsc{stale}.
& The guideline and dialogue remain at \(g_0\), so the committed
evidence remains fresh.
& A committed result cannot support satisfaction after a bound
dependency changes. \\

11
& Repair and revalidation
& Repair the local script reference, revalidate at \(v_2\), and commit
new \textsc{passed} evidence that supersedes the \(v_1\) entry.
& No repair is required; options B--D remain unsupported because they
contradict the active rule and observed dialogue.
& Repair changes current state; it does not revive stale evidence. \\

12
& Finalization gate
& \(o_S\) is satisfied only if its evidence is fresh and its
satisfaction predicate holds; other active obligations are checked
separately.
& \(o_G\) is satisfied only if its evidence is fresh and its rule-local
satisfaction predicate holds.
& Satisfaction of one obligation is necessary but not sufficient for
global finalization. \\

\bottomrule
\end{tabularx}

\caption{Representative obligation-level protocol replays embedded in
the end-to-end SkillsBench and GuideBench workflows. Each row
illustrates an authority boundary for the selected obligation; other
active obligations are omitted for space. A committed
\textsc{passed} result contributes authoritative evidence, while
satisfaction additionally requires the obligation predicate and current
freshness. Finalization requires satisfaction of all active mandatory
obligations. The examples use actual agent-visible benchmark materials,
and the SkillsBench dependency mutation is controlled.}
\label{tab:detailed-execution-traces}
\end{table*}

For each selected obligation, the replay traverses the same
state-transition discipline:

\begin{equation}
\begin{aligned}
\text{source}
&\rightarrow \text{qualified obligation} \\
&\rightarrow \text{scoped proposal}
\rightarrow \text{observed evidence} \\
&\rightarrow \text{versioned commitment}
\rightarrow \text{freshness check} \\
&\rightarrow \text{global finalization check}.
\end{aligned}
\label{eq:execution-trace}
\end{equation}

The SkillsBench replay shows how one artifact obligation is authorized,
observed, committed, invalidated, and revalidated. It does not enumerate
the task's other obligations concerning copied data, rendered content,
layout, interaction, or visual behavior. The agent's file writes remain
proposals until trusted workspace observation produces admissible
evidence. Changing a bound artifact invalidates the earlier commitment,
and repair requires new evidence at the current dependency version.

The GuideBench replay follows \texttt{rule\_6} and its supported
decision while omitting unrelated or additional active rule-local
obligations. The proposed option does not establish the answer state.
The bound validator checks both applicability and decision agreement for
the same obligation before committing authoritative evidence.

These representative obligation-level replays demonstrate the shared
authority discipline without treating a committed \textsc{passed}
result as automatic satisfaction or treating satisfaction of the
selected obligation as sufficient for task completion. In a complete
run, an obligation is satisfied only when its commitment predicate holds
and its evidence is fresh. The finalization gate ranges over the full
active mandatory set and permits completion only when every member is
satisfied.

\end{document}